\pdfoutput=1

\documentclass{article}

\PassOptionsToPackage{numbers}{natbib}
\usepackage[final]{AdvML_Frontiers_2024}

\usepackage[utf8]{inputenc} % allow utf-8 input
\usepackage[T1]{fontenc}    % use 8-bit T1 fonts
\usepackage{hyperref}       % hyperlinks
\usepackage{url}            % simple URL typesetting
\usepackage{booktabs}       % professional-quality tables
\usepackage{amsfonts}       % blackboard math symbols
\usepackage{nicefrac}       % compact symbols for 1/2, etc.
\usepackage{microtype}      % microtypography
\usepackage{xcolor}         % colors

\usepackage{multirow}  % Add this line to include the multirow package
\usepackage{adjustbox}
\usepackage{subcaption}

\title{Learning to Forget using Hypernetworks}

\usepackage{amsthm}
\usepackage{amsmath}
\usepackage{algorithm,algpseudocode}

\newtheoremstyle{customdef} % Custom style for definitions
{10pt} % Space above
{10pt} % Space below
{\itshape} % Body font (italicized)
{} % Indent amount
{\bfseries} % Theorem head font (bold)
{.} % Punctuation after theorem head
{ } % Space after theorem head
{\thmname{#1} \thmnumber{#2} \thmnote{#3}} % Theorem head spec

\theoremstyle{customdef}
\newtheorem{definition}{Definition}[section] % Numbered within section

\newcommand\affiliation[1]{\textsuperscript{\textnormal{#1}}}

\newcommand\cam{1}
\newcommand\mila{2}

\author{%
    Jose Miguel Lara Rangel%
    \affiliation{\cam}
\quad
    Stefan Schoepf%
    \affiliation{\cam}
\quad
    Jack Foster%
    \affiliation{\cam}
\\[1ex]\bf
    David Krueger%
    \affiliation{\mila}
\quad
    Usman Anwar%
    \affiliation{\cam}
}

\newcommand\thanksline{%
    \affiliation{\cam}University of Cambridge.
    \affiliation{\mila}MILA \& Université de Montréal.
}

\makeatletter
\let\ognotice\@noticestring
\renewcommand\@noticestring{\thanksline\par\smallskip\ognotice}
\if@preprint\else\if@neuripsfinal\else\renewcommand\@noticestring\ognotice\fi\fi
\makeatother

\begin{document}

\maketitle

\begin{abstract}

   Machine unlearning is gaining increasing attention as a way to remove adversarial data poisoning attacks from already trained models and to comply with privacy and AI regulations. The objective is to unlearn the effect of undesired data from a trained model while maintaining performance on the remaining data. This paper introduces HyperForget, a novel machine unlearning framework that leverages hypernetworks -- neural networks that generate parameters for other networks -- to dynamically sample models that lack knowledge of targeted data while preserving essential capabilities. Leveraging diffusion models, we implement two Diffusion HyperForget Networks and used them to sample unlearned models in Proof-of-Concept experiments. The unlearned models obtained zero accuracy on the forget set, while preserving good accuracy on the retain sets, highlighting the potential of HyperForget for dynamic targeted data removal and a promising direction for developing adaptive machine unlearning algorithms. 
  
\end{abstract}

\section{Introduction}
\label{intro}

The need for machine learning (ML) models to forget specific points of their training data has become an essential requirement due to increasing security, ethical, and regulatory concerns in AI. Data forgetting is a critical defense mechanism against adversarial attacks that manipulate models to change their behavior or extract training data information from them \citep{goel2024corrective, schoepf2024potion, carlini2024poisoning, ren2020adversarial, marchant2022hard, cao2015towards, Nasr2023ScalableEO}.
Additionally, regulations like the GDPR with its Right-to-be-Forgotten (RTBF) or the EU AI Act enhance individuals' data privacy rights, allowing them to request the deletion of their data \citep{voigt2017eu, GDPRSins, eu_ai_act, zhang2023right}.
As ML models capture information of their training data in their parameters, aligning them with ethical and regulatory standards requires not only to delete stored data but also to remove its influence on the parameters, which directly impacts the model’s capabilities \citep{Bourtoule2021MachineUnlearning, foster2024fast}.

Machine unlearning (MU) focuses on developing algorithms capable of efficiently and effectively removing the influence of specific data on an ML model, while maintaining unrelated knowledge or capabilities unaffected \citep{baumhauer2022machine,liu2024rethinking, Bourtoule2021MachineUnlearning}. Ideally, an unlearned model should behave identically to a model that was never trained on the data being unlearned in the first place. Thus, for randomly initialized models, exact unlearning is achieved when the distribution of unlearned models is identical to the distribution of models trained on the dataset $D$ excluding the forget set $D_f \subseteq D$, either by equating their distribution of parameters or outputs \citep{brophy2021machine, ginart2019making, nguyen2022survey, kurmanji2024towards, Thudi2021UnrollingSU}. The gold standard for exact unlearning is retraining the model from scratch without the forget set, which is costly in time and resources as it requires full access to the training dataset and must be done for each forgetting request.

Consequently, research has focused on the development of approximate unlearning methods to mitigate the retraining drawbacks  \citep{graves2021amnesiac, chundawat2023zero, liu2024rethinking, nguyen2022survey, sekhari2021remember}. An ideal unlearning algorithm should be consistent with the retrained model outputs, preserve as much performance as possible on the retain set $D_r=D \backslash D_f$, be faster than retraining, provide guarantees of effective removal of $D_f$ influence, be lightweight, scalable, and avoid recovering unlearned data in strict compliance scenarios. Achieving all these conditions is challenging and often involves trade-offs based on application needs, forget set size, and available resources \citep{nguyen2022survey, zhang2023review, ginart2019making, tarun2023fast, guo2019certified, yoon2022few, micaelli2019zero, golatkar2020eternal, he2021deepobliviate, shuang2024forgetting}.

Pre-trained methods are proactive and embed unlearning capabilities into the model’s design or training process, which makes them efficient, robust and reliable at removing undesired data points, but require complex designs and higher computational overhead during training \citep{Bourtoule2021MachineUnlearning, wang2024networkdiffusion,zhang2023right}. Post-training methods, the most popular type of unlearning methods, are reactive and can be applied on existing models without redesign, but may not always fully remove data influence \citep{chundawat2023zero, foster2024informationtheoreticapproachmachine, goel2024corrective}.

Also, unlearning algorithms can be model intrinsic, agnostic, or involve data modifications for faster retraining; and may need full, partial, or no access to training data \citep{Bourtoule2021MachineUnlearning, chundawat2023zero, tarun2023fast}. Particularly, we refer as Retrieval-Enhanced Unlearning (REU) to methods that use stored metadata or auxiliary data saved during the model's training process, distinct from the actual training data, to induce unlearning-- inspired by human Retrieval-Induced Forgetting phenomena \citep{BIOFORGETTING, murayama2014forgetting}. A relevant concern is scalability due to potential challenges in metadata storage and retrieval \citep{chundawat2023zero, liu2024rethinking, nguyen2022survey, thudi2022unrolling}.

We propose a new REU pre-trained framework that treats forgetting as a generative process with a sequence of states with varying levels of forgetting. This process transitions from less to more forgetting, positioning forgetting as the inverse of learning—an unlearning process. By reconstructing these states, we aim to place the model in a state where it has gained essential knowledge on the retain set, but not on the forget set. For this, we employ hypernetworks to generate parameters with reduced performance on the forget set, while preserving performance on the retain set. We name this approach HyperForget and, to the best of our knowledge, it is the first use of hypernetworks for MU.

\begin{figure}
    \centering
    % First subfigure
    \begin{subfigure}[b]{0.56\textwidth} 
        \centering
        \includegraphics[width=\textwidth]{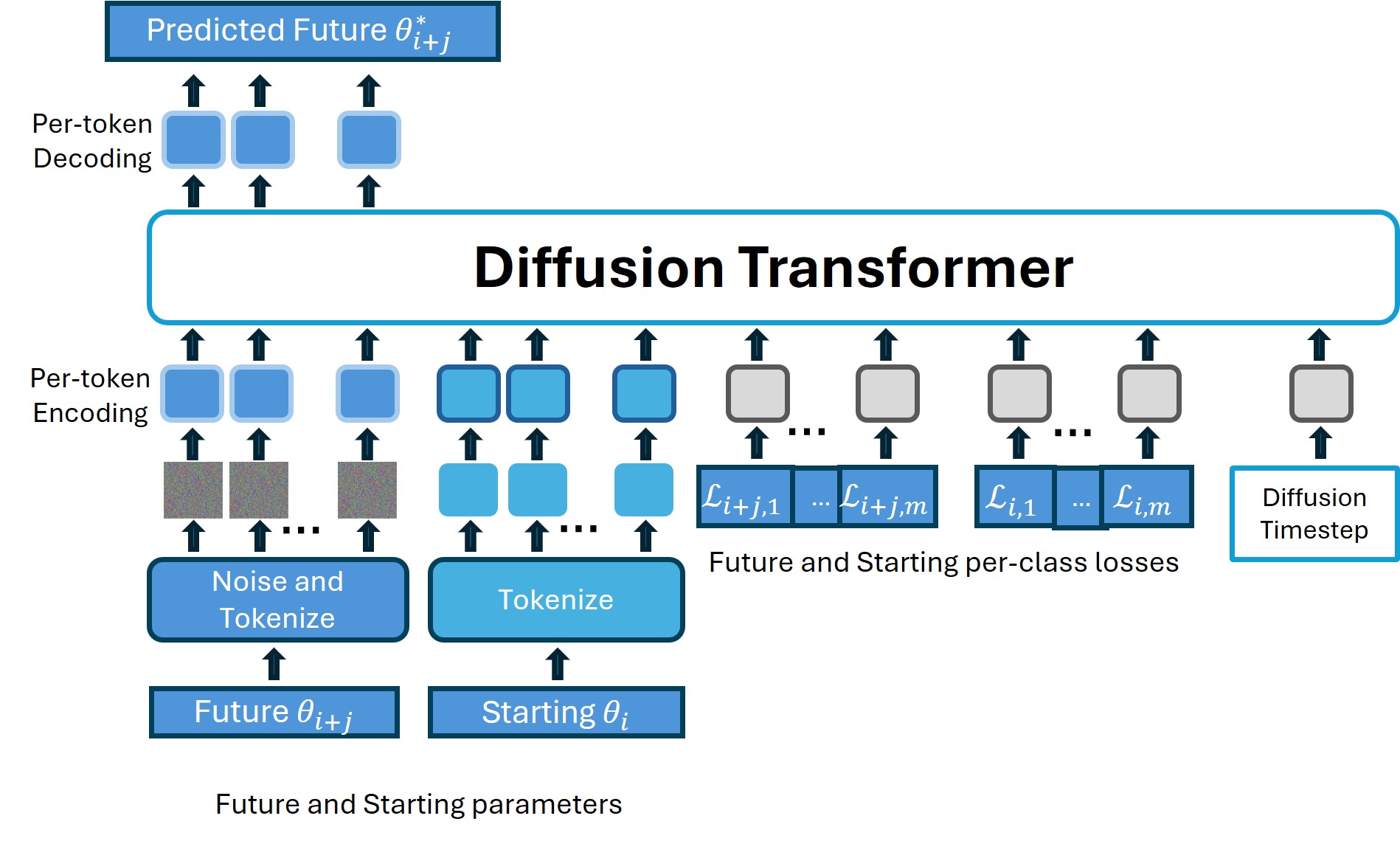}
        \caption{DiHyFo-1}  
        \label{fig:dyhifo1_full}
    \end{subfigure}
    \hfill
    % Second subfigure
    \begin{subfigure}[b]{0.37\textwidth} 
        \centering
        \includegraphics[width=\textwidth]{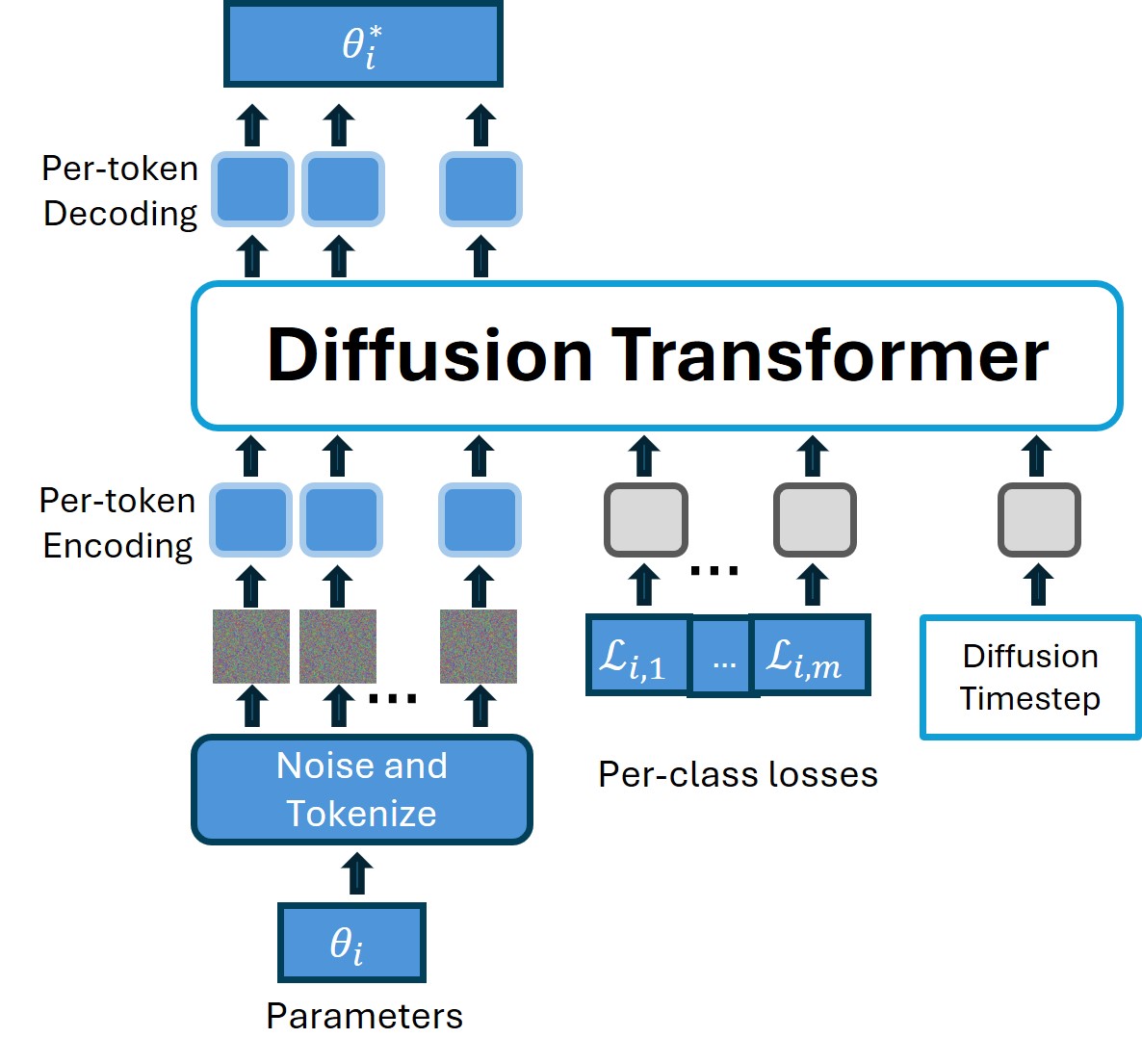}
        \caption{DiHyFo-2}  
        \label{fig:dyhifo2_full}
    \end{subfigure}

    \caption{Two implementations of Diffusion Hyperforget Networks. DiHyFo-1 is conditioned on current parameters, current losses and target losses to work as a learned optimizer-deoptimizer, while DiHyFo-2 is conditioned only on target losses to work as a conditioned generator.}  
    \label{fig:dihyfos}
    \vspace{-1em} 
\end{figure}

By integrating diffusion models as hypernetworks \citep{sohl2015deep, ho2020denoising, song2020score}, we create two Diffusion HyperForget Networks (DiHyFo) shown in Figure~\ref{fig:dihyfos}, and use them to sample unlearned models for MU tasks in Proof-of-Concept (POCs) scenarios, comparing them with models retrained from scratch without the forget sets to show their potential for MU. Notably, while a retrained model should be constructed for each forget set, one DiHyFo can sample unlearned models for all the forget sets and can be interpreted and evaluated using both the parameter and the output space perspectives of unlearning. Our results suggest that the sampled unlearned models effectively achieve zero performance on forget sets while maintaining high accuracy on retain sets, and closely mimic a retrained model. Our contributions are:

\begin{enumerate}
    \item We conceptualize and approach forgetting for ML models as a generative process, making a model to forget by positioning it in a state where its parameters have gained relevant knowledge and capabilities on the retain set but not on the forget set.
    
    \item We propose a novel REU framework leveraging hypernetworks and diffusion models. We present two implementations and POCs to show their potential for unlearning, and highlight significant limitations and potential future directions.
    
    \item One DiHyFo can sample unlearned models for different forget set configurations, enabling the development of more adaptable unlearning methods for dynamic forgetting requests.
    
\end{enumerate}

\section{Related work}
\label{rel_work}

\paragraph{Hypernetworks.} A neural network $G(C; \theta_G)$ with learnable parameters $\theta_G$ and context input $C$ is a hypernetwork if its output are parameters for another network $F(X;\theta_F)$ that solves a task $T$ with associated data $D=\{X,Y\}$, i.e. $\theta_F=G(C; \theta_G)$. Therefore, in this framework $\theta_G$ are optimized to use $G$ to sample parameters $\theta_F^*$ that can be used by $F$ to process $X$ and predict $Y^*$ for task $T$ \citep{chauhan2023brief, ha2016hypernetworks}. 

This approach can reduce the parameter search space and offers adaptability as, unlike conventional parameters that remain fixed after training, it can be conditioned to sample parameters for multiple related tasks \citep{ha2016hypernetworks, mahabadi2021parameter, oh2022cvae}. It has shown promise in continual learning, transfer learning, weight pruning, and uncertainty-aware modeling \citep{von2019continual, chauhan2023brief, volk2022example, li2020dhp, kristiadi2019predictive}. However, challenges limit broader adoption, such as scalability, lack of understanding of representational capacity, learning dynamics, generalization, and ensuring parameters are valid and not merely memorized or interpolated  \citep{ chauhan2023brief, li2021heterogeneity, li2020dhp}.

\paragraph{Diffusion Hypernetworks.} Similar to diffusion models \citep{sohl2015deep, ho2020denoising}, neural networks training with SGD transforms randomly initialized parameters (noise) into a structured set that forms a specific distribution. This inspired the use of diffusion as hypernetworks, what we call Diffusion Hypernetworks (DiHy). While this is still an emerging area of research, early approaches like G.pt \cite{Peebles2022LearningTL} show that diffusion models have significant potential to act as hypernetworks. \footnote{Although G.pt was not categorized as hypernetwork in \citep{Peebles2022LearningTL}, mainly due to it predicting targeted parameters updates instead of direct parameters, we categorize it as a hypernetwork according to our previous definition.} G.pt acts as a learned optimizer. By using a diffusion transformer (DiT) and DDPM \citep{peebles2023scalable, ho2020denoising} it takes current parameters (potentially random noise) with their associated performance and target performance, and update them to obtain the desired performance, indirectly solving optimization problems traditionally handled by SGD. G.pt has demonstrated broader generative properties than other approaches, although it is less precise in achieving target metrics, requires big data sets constructed by saving checkpoints from multiple main network training runs, involves considerable training overhead, and shows limited capabilities to extrapolate to performances beyond its training data \cite{wang2024networkdiffusion, Peebles2022LearningTL}. However, despite these limitations, we show that its ability to learn a conditional generative model over weights of the main network is useful towards MU. We replicate some experiments related to G.pt (originally reported in \cite{Peebles2022LearningTL}) and list some observations that might be of independent interest. See Appendix \ref{subsec:gpt_observations}. 

Furthermore, \cite{erkocc2023hyperdiffusion} also used DiT and DDPM to generate unconditioned parameters, \cite{wang2024networkdiffusion} improved computational efficiency by incorporating an autoencoder and U-Net for learning on a parameters latent space. Both works used datasets with only optimized parameters, effectively introducing a soft bias for high-performing parameters generation, but limiting capabilities for more diverse parameter distributions. Recently, \cite{li2024text} extended \cite{Peebles2022LearningTL} ideas to text-to-model generation, where a customized classifier for a subset of classes of a large dataset can be sampled by prompting text.

\section{Proposed method}

We propose \textbf{HyperForget} -- a hypernetwork based framework for unlearning in which a hypernetwork is trained to learn a distribution over model parameters.
The hypernetwork is then used to sample model parameters that solve a forgetting task $U$.
Thus, the resulting model, a HyperForget Network (HyFo), can generate parameters that simultaneously yield high-performance on the retain set $D_r$ and low-performance on the forget set $D_f$ associated to task $U$. Indeed, it reconstructs the knowledge and capabilities for $D_r$ while effectively forgetting $D_f$, achieving this efficiently in a single forward pass, eliminating gradient calculations typically required by other unlearning methods, and allowing for dynamic unlearning adaptation in scenarios with frequent and varying forget requests.

In this work, we use DiT as the hypernetwork. We call the resulting model Diffusion HyperForget Network (DiHyFo). Integrating diffusion models into the HyperForget framework provides a structured approach to gradual unlearning and allows to control the level of data influence removal. We explored two options to construct a DiHyFo for class-level unlearning.

Consider a classification task $T$ with a dataset $D$ formed by examples $x_i \in X$ with associated labels $y_i \in \{1,...,m\}, m\in \mathbb{N}$, drawn from an unknown distribution $P$.
A model $F$ trained to solve $T$ is requested to forget a subset of classes, i.e., $D_f$ contains data from specific class labels.
For this task, we employ a DiHyFo conditioned on class losses to generate parameters that obtain high performance in $D_r$ and low performance in $D_f$.

\textbf{DiHyFo-1 }uses a DiT and follows \citep{Peebles2022LearningTL} but conditioned at a class-level. 
% In this scenario, parameter adjustments impact multiple classes simultaneously. 
This allows us to control losses for specific classes at the same time, e.g.,
prompt DiT to generate parameters with high loss values on classes in the forget set, but low loss values on classes in the retain set.
This is a different and more challenging optimization problem than the one solved in \citep{Peebles2022LearningTL}.
% different to \citep{Peebles2022LearningTL} that constructs a learned optimizer to update parameters with decreased losses, we train DiHyFo-1 to learn bidirectional loss movements, i.e., it learns to learn and to forget. 

During training, current parameters $\theta$, future parameters $\theta^*$, and their associated class losses $\mathcal{L}_1, ..., \mathcal{L}_m$ and $\mathcal{L}_1^*, ..., \mathcal{L}_m^*$ for task $T$ are randomly selected from a run uniformly sampled from a dataset of checkpoints grouped by runs. $\theta$ is always from an earlier step than $\theta^*$. Parameters are normalized and tokenized layer-by-layer flattening them into 1D vectors, each corresponding to a unique token. Layers with both weight and bias are decomposed into separate tokens, and each token accepts a maximum number of parameters set to be smaller than the DiT hidden size. 

The time-step, losses, and their differences $\Delta_{\mathcal{L}_1, \mathcal{L}_1^*},...,\Delta_{\mathcal{L}_m, \mathcal{L}_m^*}$ are tokenized with a frequency-based encoder. All the tokens are linearly projected to the DiT hidden size, and Standard Gaussian noise is added to the future parameters at each time-step, and then passed to the DiT to learn to denoise them. The DiT has a decoder at the end that linearly projects each token back to its original size, and a residual connection to predict the parameter updates $\theta_t^*-\theta_t$. The training objective is to minimize the MSE between the generated parameters and the original future parameters. This enables to sample parameters directly in the parameter space with the desired losses for each class.

At inference, the model takes a set of current parameters (potentially random noise) with current and target losses for each class, and returns the updated parameters for the requested losses by denoising an input Gaussian noise vector via DDPM with fixed variances \citep{ho2020denoising}. Thus, DiHyFo-1's goal is to predict the distribution of updated parameters that achieve each target class loss. 

\textbf{DiHyFo-2 }is directly conditioned on the desired class losses, Figure \ref{fig:dyhifo2_full}, which has shown good results in other tasks using DiHy \citep{li2024text, wang2024networkdiffusion}. The DiT is conditioned on a large set of examples of parameters with both high and low performance on individual classes, enabling to synthesize new parameters tailored to specific losses directly in the parameter space. Thus, the task of DiHyFo-2 is to predict the distribution of parameters that achieve the desired losses.%, Equation \ref{eq:dihyfo2_gen_task}.

Both models use a DiT with hidden dimension of 1536, 12 hidden layers, and 16 attention heads, trained with AdamW \citep{loshchilov2017decoupled} maintaining an exponential moving average of the model weights across the training process. Learned positional embeddings, initialized to zero, are applied across all tokens. Additional details of the framework and both models can be found in Appendix \ref{subsec:add_hyperforget}.

\begin{figure}
    \centering
    % Second row, first subfigure
    \begin{subfigure}[b]{0.48\textwidth}
        \centering
        \includegraphics[width=\textwidth]{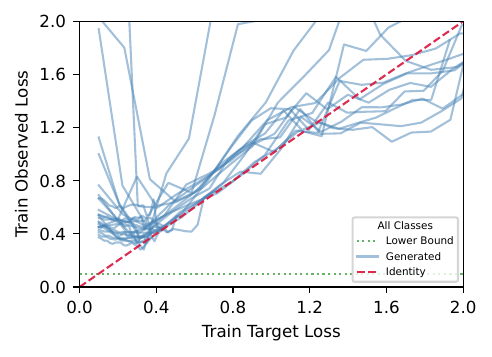}
        \caption{Dihyfo-1 obtained losses for class 2 of MNIST} 
        \label{fig:dihyfo2_mnist4_obs_test0}
    \end{subfigure}
    \hfill
    \begin{subfigure}[b]{0.48\textwidth}
        \centering
        \includegraphics[width=\textwidth]{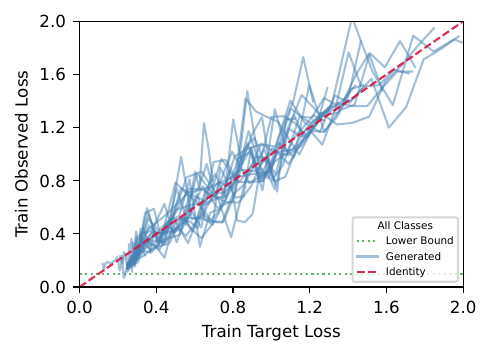}
        \caption{Dihyfo-2 obtained losses for class 2 of MNIST-4} 
        \label{fig:dihyfo2_mnist4_obs_train2}
    \end{subfigure}
    
    \caption{Examples of observed vs target losses for class 2 using models sampled with DiHyFo.}
\label{fig:pobs_des_dihyfo2_mnist4}
\end{figure}

\section{Experimental Setup}
\paragraph{Checkpoints datasets.} To reduce the complexity of the task, we consider a simplified scenario with $m$ classes, out of which $r$ classes are considered as `pivot' classes.
For pivot classes, we only consider high-performing parameters.
For the remaining $m-r$ class losses, DiHyFo models are trained on parameters with varied values of losses.
In general, any data that a model designer is certain would not need to be forgotten later can act as a pivot, simplifying the learning problem.
The forget set can be any subset of the $m-r$ classes, while the retain sets must always include the pivots.
More complex scenarios can consider a small number of pivots or no pivots. 

Checkpoints of parameters and their class losses are collected during multiple training runs of an MLP classifier trained on MNIST. A random subset of classes is selected for undersampling in each run to capture more diverse losses, and during training checkpoints are randomly selected and evaluated for saving. Similar to prior work \citep{Peebles2022LearningTL,roeder2021linear, schurholt2021self}, we apply permutation augmentation to help DiHyFo learn that weights of neural networks are permutation invariant.
To provide DiHyFo-1 with examples of losses increments we delete a random subset of classes during MLP training in some runs and save the corresponding checkpoints. 
We further use heuristics to track the evolution of class losses and prioritize checkpoints with varied losses across classes over checkpoints with similar losses across different classes.
For a detailed description of the checkpoint datasets collection see Appendix \ref{subsec:datasets_generation}.

For our experiments, we consider two variations of MNIST dataset -- full MNIST and MNIST-4 with four classes.
For MNIST-4, we use classes \{0,1\} as pivots. We save 150 checkpoints during training runs of 25 epochs of a two-layer MLP with two hidden units and ReLU activation, totaling 1.9M checkpoints. For full MNIST, we use  \{5-9\} as pivot classes, seven hidden units MLP, and 200 checkpoints saved per training run for a total of 3.7M checkpoints.

\paragraph{Evaluation procedures.} The ability of DiHyFo to generate class-loss targeted parameters is evaluated by analyzing losses on train and test set. Following \cite{Peebles2022LearningTL}, we use the metric of prompt alignment (see Appendix~\ref{subsec:evaluation_metrics}), which intuitively measures the alignment between the observed losses and target losses. A prompt alignment score of 1 indicates perfect alignment. Further, we include the correlation between both losses as a complementary metric, and plot them along the identity line with a reference of high performance—often the median or average of losses in checkpoints.

As there are no general frameworks or benchmarks in MU, a typical strategy to assess the performance of an MU method is to compare the unlearned model with a model retrained from scratch only on the retain set. The closer the behavior of the unlearned model to the retrained model, the more effective the unlearning. Thus, to assess unlearning capabilities of DiHyFo, we sample unlearned models and compare against a retrained model. We consider \{2\} and \{2,3\} as forget sets for MNIST-4, while \{2\}, \{2,3,4\}, and \{0,1,2,3,4\} for MNIST. Notably, each forget set required a retraining process, whereas a single DiHyFo model can sample unlearned models for all forget sets, highlighting its suitability for scenarios with dynamic unlearning requests. We compare accuracy on test sets and member inference attacks score (MIA) to assess the unlearning consistency and guarantees against adversarial attacks to infer forget sets information \cite{chundawat2023can}. The output and parameter spaces of the sampled unlearned models are compared against the retrained model by computing the predictions overlap, and the \textit{unlearning score}. Unlearning score is the Jensen-Shannon Divergence (JSD) between the probability distributions outputted by the sampled unlearned model and a candidate retrained model \citep{foster2024informationtheoreticapproachmachine, nguyen2022survey, chundawat2023can}. As we only consider class-forgetting setting, ideally, the predictions overlap on the retain set should be high between the unlearned model and retrained model, while for the forget set a low overlap would be more desirable. 
Similarly, a high unlearning score value is preferred as this indicates that the unlearned model's predictions are similar to those of the retrained model. The maximum value of the unlearning score can be $1$ and the lowest value can be $0$.

\section{Results and Discussion}

\paragraph{Generative performance.} 
We measure the  capabilities of each model to generate parameters with the desired loss by computing the prompt alignment \citep{Peebles2022LearningTL} and correlation between the target loss and the actual loss obtained with the sampled parameters.
Both these metrics show initial fluctuations but tend to increase and stabilize trough training. As depicted in Figure \ref{fig:pobs_des_dihyfo2_mnist4}, although the observed losses generally align, it is challenging for the model to precisely match the target losses, particularly for higher loss values (see Appendix \ref{subsec:training} for additional results). 
Similarly, the observed losses for pivots track the target losses, successfully retaining performance but struggling with high losses and jumps between levels, as expected with the dataset structure.
As we focus on low-loss regions for pivots, these deviations are less relevant for our evaluations.  

Learning bidirectional loss movements at a class level is a challenging task, leading to less precise prompt alignment than in other tasks \citep{Peebles2022LearningTL}. However, this is not a major issue as unlearning typically focuses on high and low-performing parameters rather than covering the entire loss range.
While for MNIST-4 DiHyFo-2 showed to be better at aligning losses, for MNIST DiHyFo-1 showed better consistency and robustness. See Appendix \ref{subsec:training} for further details.

\paragraph{Unlearning Performance.} Table~\ref{table:inv_metrics_mnist} shows that all the sampled unlearned models achieved zero accuracy on the forget sets and maintain a good accuracy on the retain sets, comparable to the retrained models.
The obtained MIA scores for all sampled unlearned models are very close to the retrained models, indicating good unlearning and robustness against inference attacks.
As shown in Table~\ref{table:comp_metrics_mnist}, the comparison of the outputs produced by the sampled unlearned models and the retrained models shows similar predictions on the retain and test sets, aligning with individual performance. Note that the perfect overlap is not expected, as our unlearned model only needs to behave like any retrained model that results from the stochastic nature of the training process, not like one specific retrained model.
% \textcolor{blue}{The lower overlap on the forget sets indicate different representations encoded in neurons for them across models.}
Moreover, the comparisons of the parameters space show high unlearning scores. Thus, the sampled unlearned models are mimicking the behavior of a retrained model, with DiHyFo-1 performing more consistently across unlearning tasks, particularly with MNIST.
See Appendix \ref{subsec:evaluation} for additional results and further discussion.

Overall, the sampled models are effectively unlearning the forget set and no longer rely on the associated data to make predictions, and preserve performance on retain sets without significant signs of catastrophic unlearning \citep{wang2024machine, nguyen2022survey}, closely approximating the unlearning gold standard. Therefore, we have a potential REU method that is model-intrinsic, accuracy-preserving with a good level of completeness, unlearning guarantees, and a potential choice for dynamic unlearning.

\begin{table}%[b]
\centering
\caption{Individual unlearned models performance on MNIST under diverse forgetting scenarios. Average over 5 sampled unlearned models.}
\vspace{0.9em}
\centering
\label{table:inv_metrics_mnist}
\adjustbox{width=1\textwidth}{%
\begin{tabular}{l c c c c c c c c c}
\toprule
\multirow{2}{*}{MNIST} & \multicolumn{3}{c}{Sampled with DiHyfo-1} & \multicolumn{3}{c}{Sampled with DiHyfo-2} & \multicolumn{3}{c}{Retrained Model} \\ 
\cmidrule(lr){2-4} \cmidrule(lr){5-7} \cmidrule(lr){8-10}
 & $D_f$=\{2\} & $D_f$=\{2,3,4\} & $D_f$=\{0,1,2,3,4\} & $D_f$=\{2\} & $D_f$=\{2,3,4\} & $D_f$=\{0,1,2,3,4\} & $D_f$=\{2\} & $D_f$=\{2,3,4\} & $D_f$=\{0,1,2,3,4\} \\
\midrule
Accuracy on $D_r$ \((\uparrow)\) & 0.9072 & 0.9390 & 0.9373 & 0.7369 & 0.9433 & 0.9518 & 0.9193 & 0.9479 & 0.9504 \\
Accuracy on $D_f$ \((\downarrow)\) & 0.0000 & 0.0000 & 0.0000 & 0.0000 & 0.0000 & 0.0000 & 0.0000 & 0.0000 & 0.0000 \\
MIA \((\downarrow)\) & 0.3682 & 0.3817 & 0.4282 & 0.3580 & 0.3243 & 0.3299 & 0.3299 & 0.3406 & 0.3798 \\
\bottomrule
\end{tabular}%
}
\vspace{-0.7em}
\end{table}

\begin{table}
\centering
\caption{Assessment of output alignment and behavioral consistency between sampled unlearned models and their corresponding retrained model on MNIST under diverse forgetting scenarios. Average over 5 sampled unlearned models.
}
\vspace{0.9em}
\centering
\label{table:comp_metrics_mnist}
\adjustbox{width=0.7\textwidth}{%
\begin{tabular}{l c c c c c c}
\toprule
\multirow{2}{*}{MNIST} & \multicolumn{3}{c}{Sampled with DiHyfo-1} & \multicolumn{3}{c}{Sampled with DiHyfo-2} \\ 
\cmidrule(lr){2-4} \cmidrule(lr){5-7}
 & $D_f$=\{2\} & $D_f$=\{2,3,4\} & $D_f$=\{0,1,2,3,4\} & $D_f$=\{2\} & $D_f$=\{2,3,4\} & $D_f$=\{0,1,2,3,4\} \\
\midrule
Outputs Overlap on $D_r$ \((\uparrow)\) & 0.9106 & 0.9511 & 0.9498 & 0.7298 & 0.9364 & 0.9482 \\
Outputs Overlap on $D_f$ \((-)\) & 0.5474 & 0.7262 & 0.7498 & 0.4221 & 0.7110 & 0.6375 \\
Outputs Overlap on Test Set \((\uparrow)\) & 0.8731 & 0.8831 & 0.8471 & 0.6978 & 0.8682 & 0.7881 \\
% Activation Distance \((\uparrow)\) & 0.1808 & 0.1738 & 0.2012 & 0.5067 & 0.1879 & 0.2953 \\
Unlearning Score ($\varphi$) \((\uparrow)\) & 0.9885 & 0.9894 & 0.9916 & 0.8478 & 0.9857 & 0.9540 \\
\bottomrule
\end{tabular}%
}
\vspace{-0.71em}
\end{table}

\section{Limitations and Future Work}
In this work, we have provided proof-of-concept experiments for a hypernetworks based approach to unlearning.
While our results indicate the potential of this approach, many opportunities for future work exist. The current formulation of HyperForget focuses on full-class unlearning, its extension and potential applicability for other unlearning tasks have yet to be explored.
A key limitation of this approach is that the generative model retains the knowledge of forget sets making this approach unsuitable for strict unlearning-for-privacy applications.
Furthermore, it is not clear from our experiments whether the proposed approach is scalable or not.
The use of diffusion models adds significant computational overhead, and it inherits concerns from DiHy models, such as limited generalization \citep{Peebles2022LearningTL, wang2024networkdiffusion}. 
Additionally, although generated parameters generally approximate target losses, it is difficult for the model to precisely match the target loss.

Future works could consider evaluating our framework on other datasets, e.g., CIFAR-10 or mini-ImageNet, as well as improving the scalability of our approach.
Future research could explore alternative architectures, improve checkpoint saving strategies, optimize the process and components, or explore other unlearning tasks. 
Improvements may come from leveraging model zoos, hypernetworks for architecture-agnostic parameter generation, and learning on latent spaces. 
Both hypernetworks and MU are in an early stage, and we hope this work inspires future research to expand these areas, contributing to more adaptive, secure, and ethical AI solutions.

%%%%%%%%%%%%%%%%%%%%END OF MAIN TEXT

% \medskip
\clearpage

{
\small

\bibliographystyle{plain}  % or another style like 'unsrt', 'alpha', etc.
\bibliography{AdvML_Frontiers_2024}  % 

}

%%%%%%%%%%%%%%%%%%%%%%%%%%%%%%%%%%%%%%%%%%%%%%%%%%%%%%%%%%%%

\newpage

\appendix

\section{Appendix / supplemental material}

This appendix provides additional information on the technical details and methods behind the proposed hypernetwork-based approach to MU and DiHyFo models. It includes explanations on data generation, experimental results, model definitions, and model evaluations.

The appendix is organized as follows:

\begin{itemize}

    \item \textbf{Section \ref{subsec:evaluation}} Additional unlearning evaluation results: This section includes further results and analysis of unlearning capabilities.

    \item \textbf{Section \ref{subsec:training}} Additional training results: We provide supplementary training results, discussing model performance, stability, and consistency across experiments to support the findings of the main study.

    \item \textbf{Section \ref{subsec:add_hyperforget}} Additional hypernetworks, HyperForget, and DiHyFo details: This section provides details on hypernetworks and how they are used for HyperForget and DiHyFo models.

    \item \textbf{Section \ref{subsec:evaluation_metrics}} Additional details on evaluation metrics: This section outlines the metrics and evaluation methods used to assess the generative performance and unlearning effectiveness of the models.

    \item \textbf{Section \ref{subsec:datasets_generation}} Additional details on datasets generation: This section describes the methodologies for dataset generation and processing, including parameter checkpoint collection and strategies for class loss tracking across different configurations.

    \item \textbf{Section \ref{subsec:gpt_observations}} Some experimental observations on G.pt: This section presents observations and insights from experiments conducted with the G.pt model, examining performance metrics and prompt alignment under various test conditions.

\end{itemize}

\subsection{Additional unlearning evaluation results}
\label{subsec:evaluation}

To apply the DiHyFo models for unlearning, we sample multiple parameters prompting different high losses values for different forget sets simultaneously with low losses for the corresponding retain sets. The generated parameters are used to load an instance of the main network, which is then evaluated on a test set. We save the sampled model that obtains the lowest average accuracy on the forget set while obtaining the highest possible average accuracy on the retain set (best unlearned model). We use accuracy for the selection and evaluation as it is a more interpretable performance measure than loss. 

\begin{figure}[H]
    \centering
    % First subfigure
    \begin{subfigure}[b]{0.496\textwidth}
        \centering
        \includegraphics[width=\textwidth]{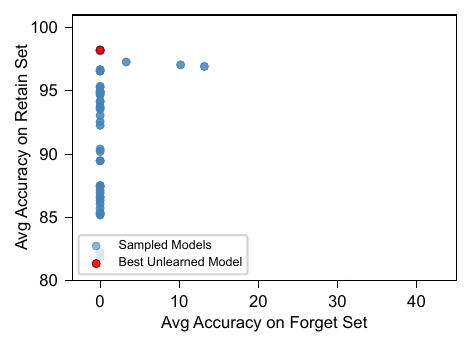}
        \caption{DiHyFo-1 sampled models, $D_f=\{2,3\}$}
        \label{fig:dihyfo1_mnist4_2}
    \end{subfigure}
    \hfill
    \begin{subfigure}[b]{0.496\textwidth}
        \centering
        \includegraphics[width=\textwidth]{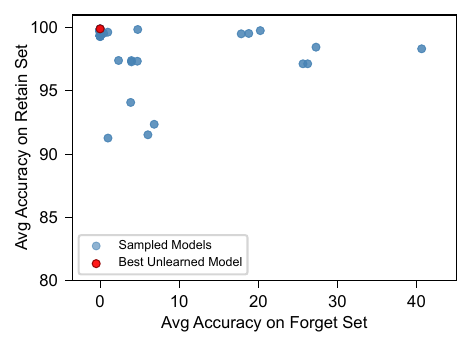}
        \caption{DiHyFo-1 sampled models, $D_f=\{2,3\}$}
        \label{fig:dihyfo1_mnist4_23}
    \end{subfigure}
    \hfill
    % Second subfigure
    \begin{subfigure}[b]{0.496\textwidth}
        \centering
        \includegraphics[width=\textwidth]{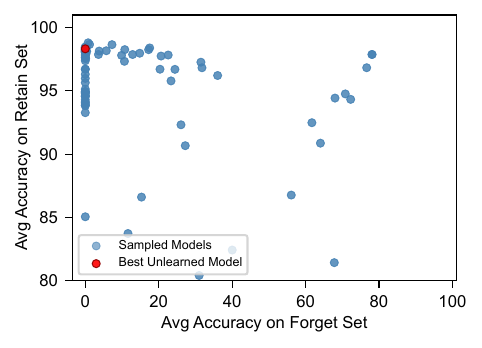}
        \caption{DiHyFo-2 sampled models, $D_f=\{2\}$}
        \label{fig:dihyfo2_mnist4_2}
    \end{subfigure}
    \hfill
    \begin{subfigure}[b]{0.496\textwidth}
        \centering
        \includegraphics[width=\textwidth]{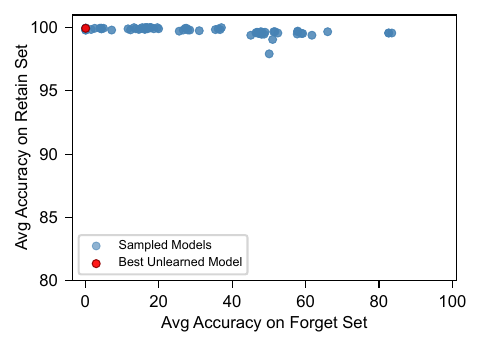}
        \caption{DiHyFo-2 sampled models, $D_f=\{2,3\}$}
        \label{fig:dihyfo2_mnist4_23}
    \end{subfigure}    
    
    \caption{Selection of unlearned models sampled using DiHyFo-1 and DiHyFo-2 on MNIST-4.}
    \label{fig:dihyfo_sampling_mnist4}
\end{figure}

\begin{figure}[H]
    \vspace{1.5in}
    \centering
    % First subfigure
    \begin{subfigure}[b]{0.496\textwidth}
        \centering
        \includegraphics[width=\textwidth]{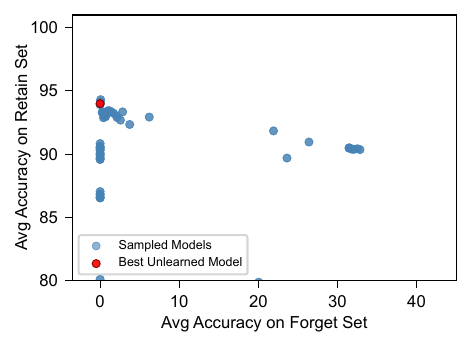}
        \caption{DiHyFo-1 sampled models, $D_f=\{2,3,4\}$}
        \label{fig:dihyfo1_mnist_234}
    \end{subfigure}
    \hfill
    \begin{subfigure}[b]{0.496\textwidth}
        \centering
        \includegraphics[width=\textwidth]{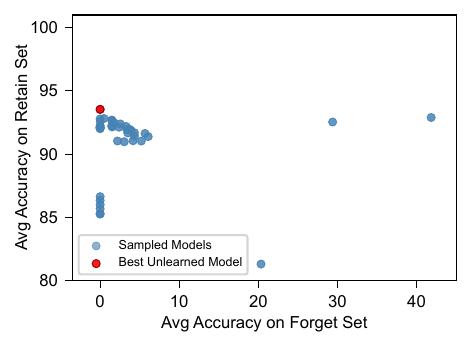}
        \caption{DiHyFo-1 sampled models, $D_f=\{0,1,2,3,4\}$}
        \label{fig:dihyfo1_mnist_01234}
    \end{subfigure}
    \hfill
    % Second subfigure
    \begin{subfigure}[b]{0.496\textwidth}
        \centering
        \includegraphics[width=\textwidth]{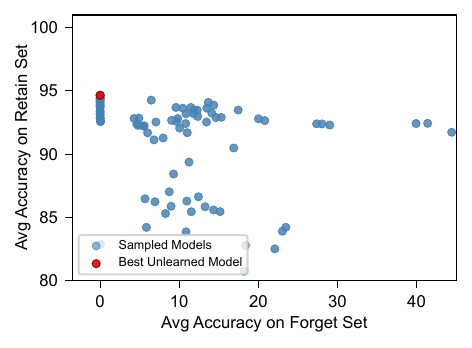}
        \caption{DiHyFo-2 sampled models, $D_f=\{2,3,4\}$}
        \label{fig:dihyfo2_mnist_234}
    \end{subfigure}
    \hfill
    \begin{subfigure}[b]{0.496\textwidth}
        \centering
        \includegraphics[width=\textwidth]{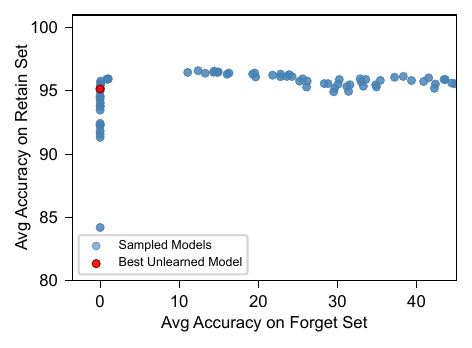}
        \caption{DiHyFo-2 sampled models, $D_f=\{0,1,2,3,4\}$}
        \label{fig:dihyfo2_mnist_01234}
    \end{subfigure}    
    
    \caption{Selection of unlearned models sampled using DiHyFo-1 and DiHyFo-2 on MNIST.}
    \label{fig:dihyfo_sampling_mnist}
\end{figure}

We sampled models that forget classes \{2\} and \{2,3\} for MNIST-4, and \{2\}, \{2,3,4\}, and \{0,1,2,3,4\} for MNIST. Figures \ref{fig:dihyfo_sampling_mnist4} and \ref{fig:dihyfo_sampling_mnist} exemplify this sampling and selection process. Tables \ref{table:inv_metrics_mnist4} and \ref{table:comp_metrics_mnist4} present the evaluation metrics on MNIST-4, showing a close behavior of the sampled unlearned models to the retrained models.

\begin{figure}[H]
\vspace{1.5in}
    \centering
    % First subfigure
    \begin{subfigure}[b]{0.49\textwidth}
        \centering
        \includegraphics[width=\textwidth]{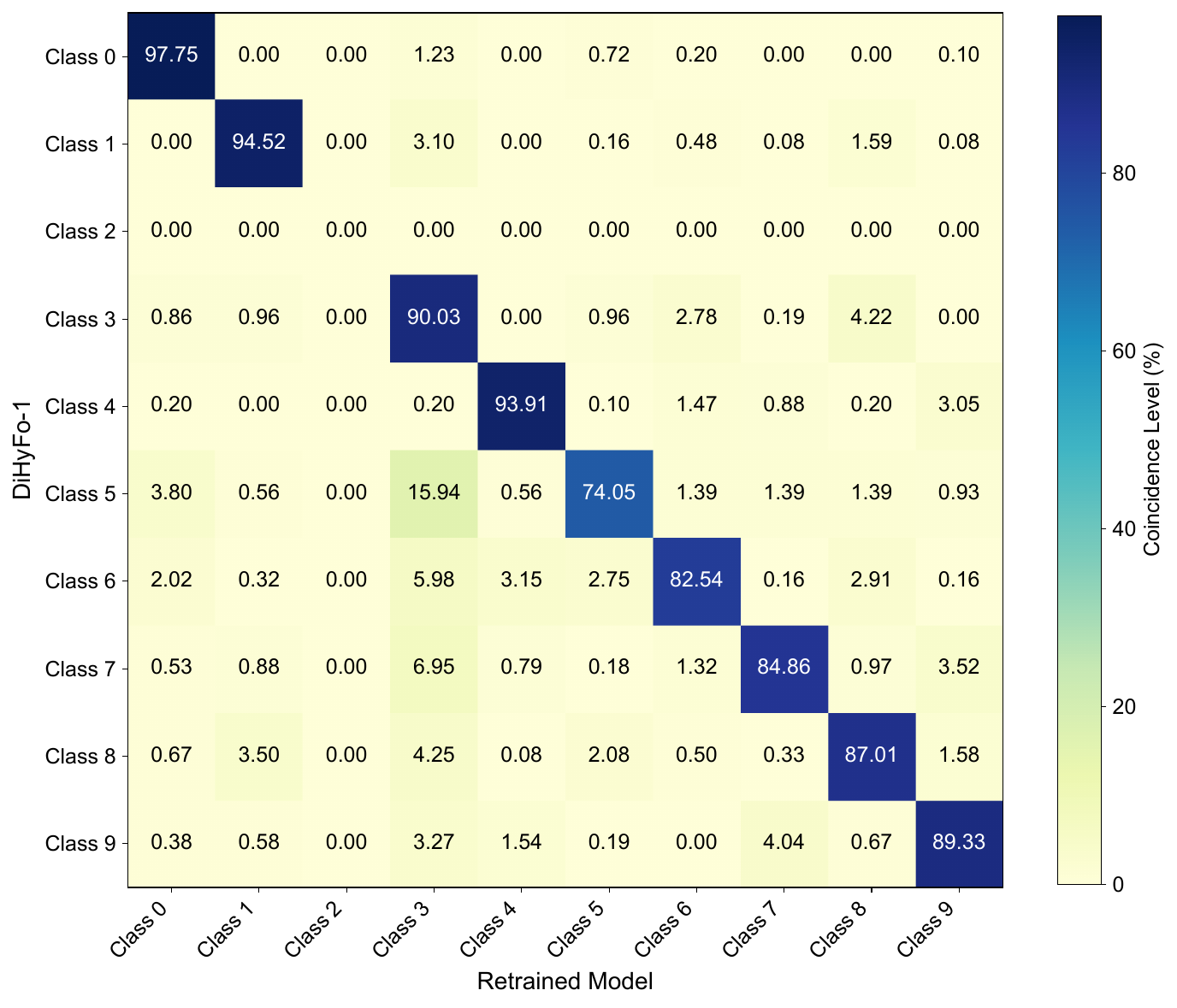}
        \caption{Predictions on test set with $D_f=\{2\}$}
        \label{fig:dihyfo1_mnist_cm_whole_2}
    \end{subfigure}
    \hfill
    % Second subfigure
    \begin{subfigure}[b]{0.49\textwidth}
        \centering
        \includegraphics[width=\textwidth]{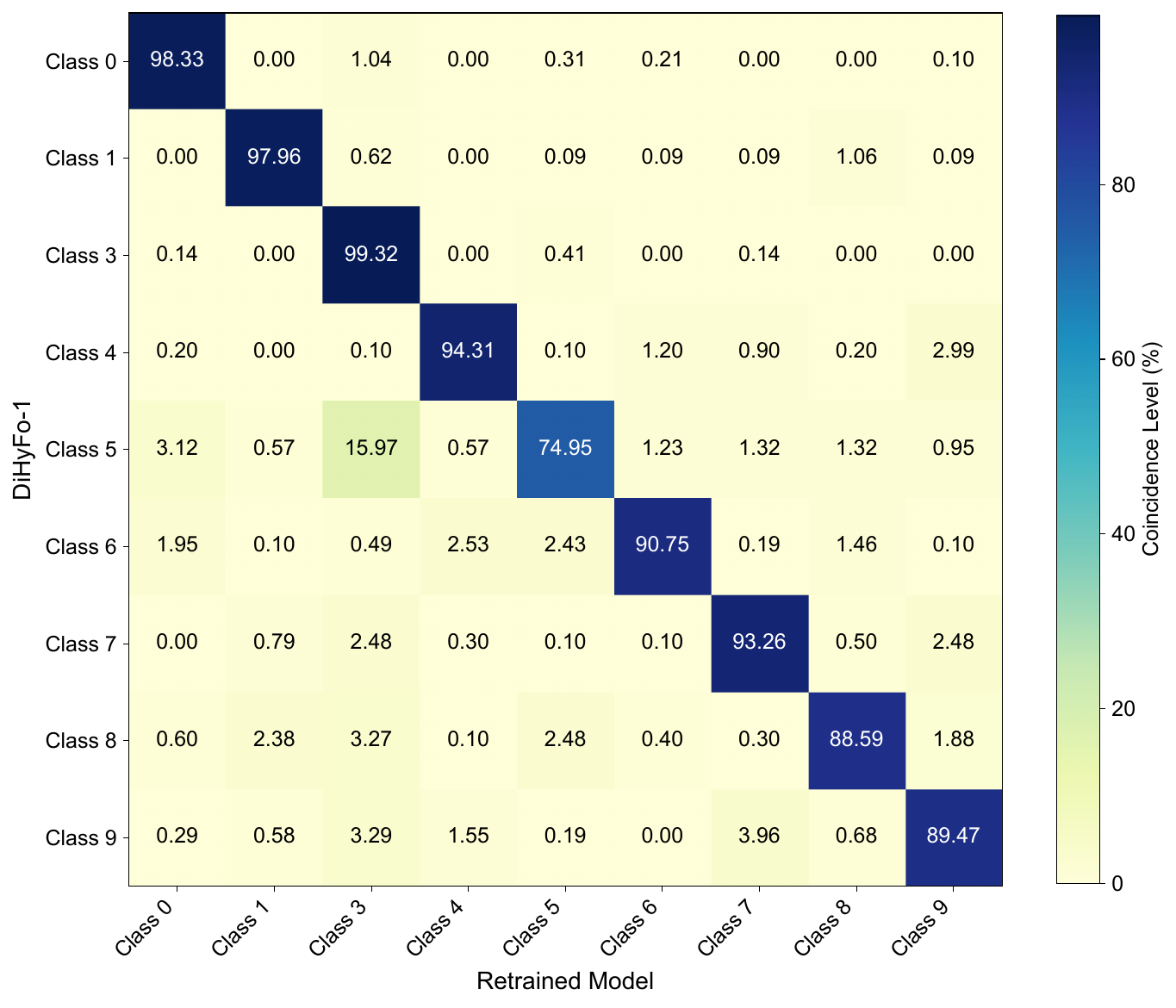}
        \caption{Predictions on $D_r$ with $D_f=\{2\}$}
        \label{fig:dihyfo1_mnist_cm_retain_2}
    \end{subfigure}

    \centering
    % First subfigure
    \begin{subfigure}[b]{0.49\textwidth}
        \centering
        \includegraphics[width=\textwidth]{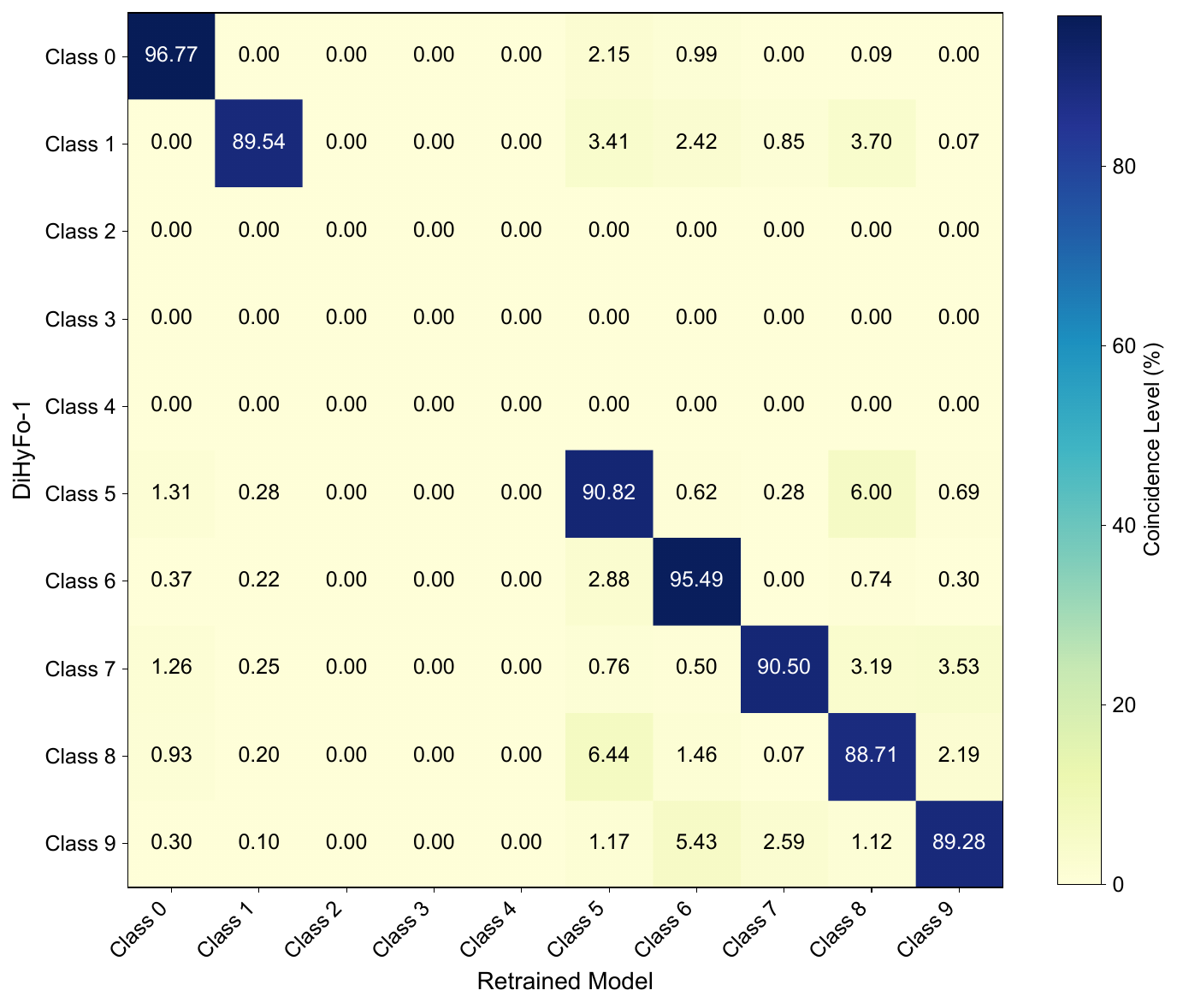}
        \caption[Confusion Matrix between DiHyFo-1 and Retrained model for $D_f=\{2,3,4\}$]{Predictions on test set with $D_f=\{2,3,4\}$}
        \label{fig:dihyfo1_mnist_cm_whole_234}
    \end{subfigure}
    \hfill
    % Second subfigure
    \begin{subfigure}[b]{0.49\textwidth}
        \centering
        \includegraphics[width=\textwidth]{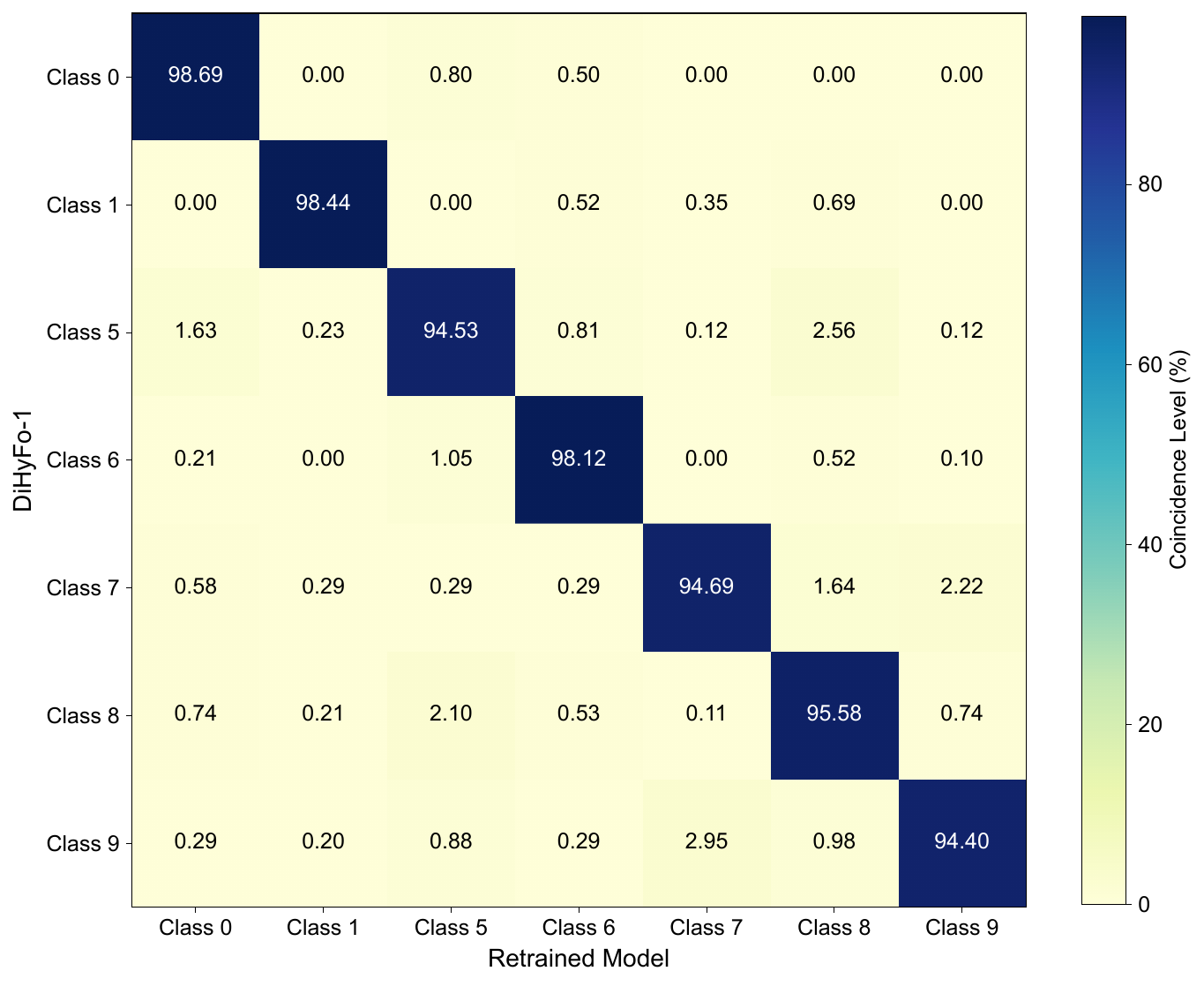}
        \caption[Confusion Matrix between DiHyFo-1 and Retrained model on $D_r$ with $D_f=\{2,3\}$]{Predictions on $D_r$ with $D_f=\{2,3,4\}$}
        \label{fig:dihyfo1_mnist_cm_retain_234}
    \end{subfigure}

    \caption{Comparison of predictions between an unlearned model sampled with DiHyFo-1 and the retrained model on MNIST.}
    \label{fig:dihyfo1_mnist_cm}
\end{figure}

\begin{figure}[H]
    \vspace{1.5in}
    \centering
    % First subfigure
    \begin{subfigure}[b]{0.49\textwidth}
        \centering
        \includegraphics[width=\textwidth]{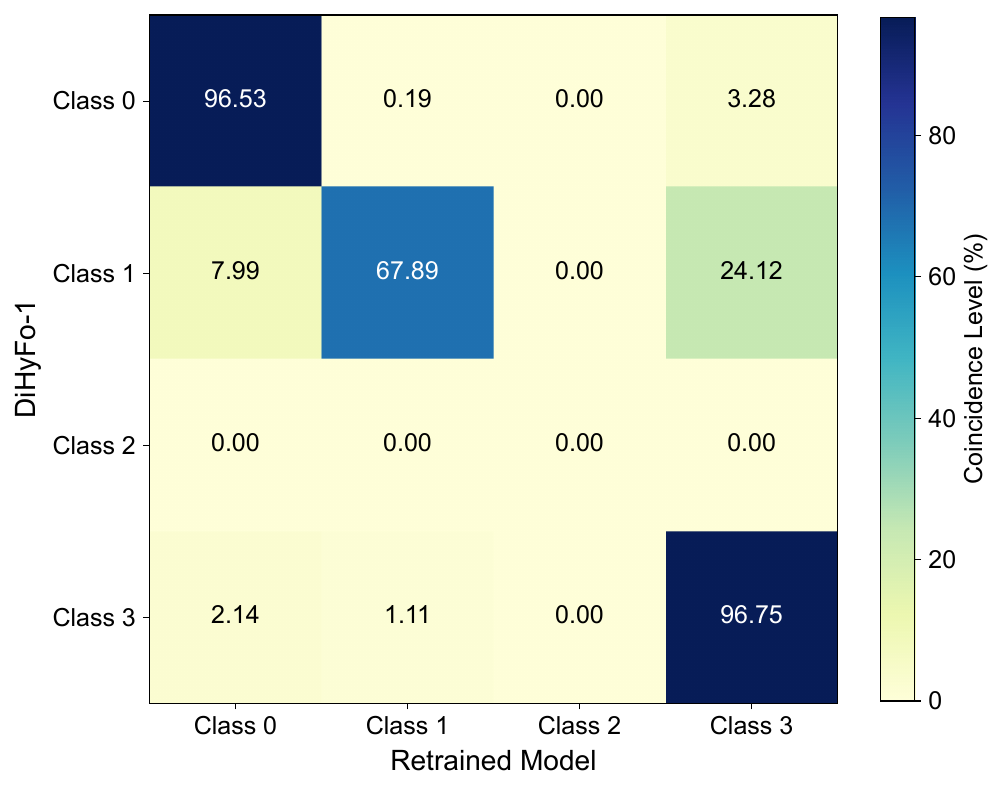}
        \caption{Predictions on test set with $D_f=\{2\}$}
        \label{fig:dihyfo1_mnist4_cm_whole_2}
    \end{subfigure}
    \hfill
    % Second subfigure
    \begin{subfigure}[b]{0.49\textwidth}
        \centering
        \includegraphics[width=\textwidth]{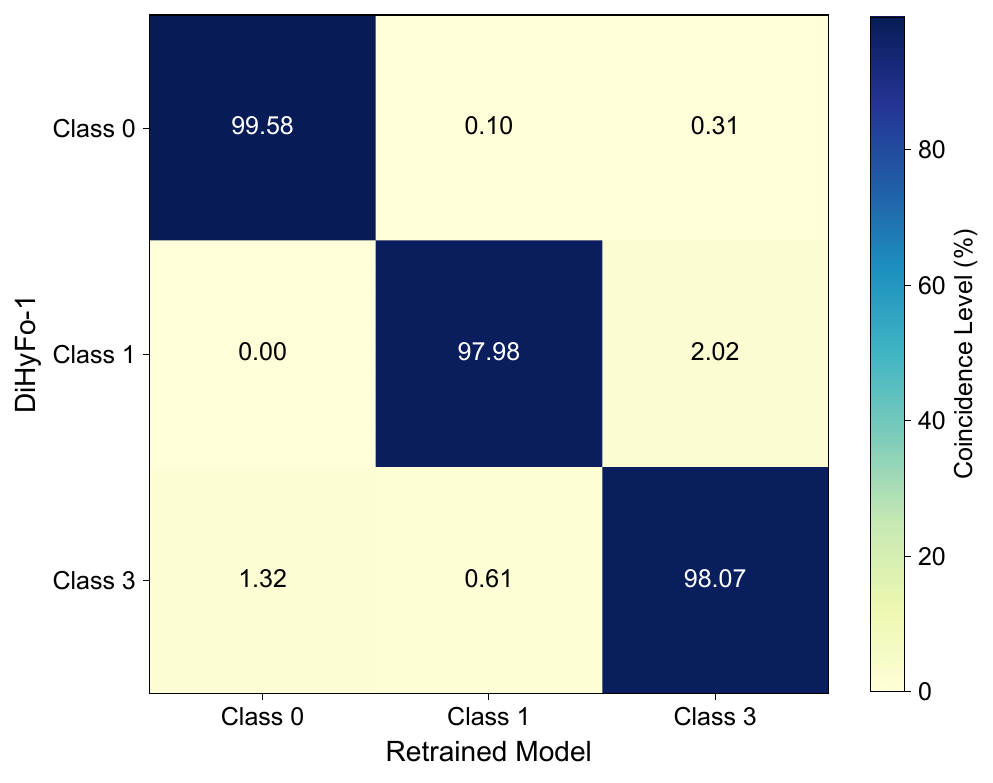}
        \caption{Predictions on $D_r$ with $D_f=\{2\}$}
        \label{fig:dihyfo1_mnist4_cm_retain_2}
    \end{subfigure}

    \centering
    % First subfigure
    \begin{subfigure}[b]{0.49\textwidth}
        \centering
        \includegraphics[width=\textwidth]{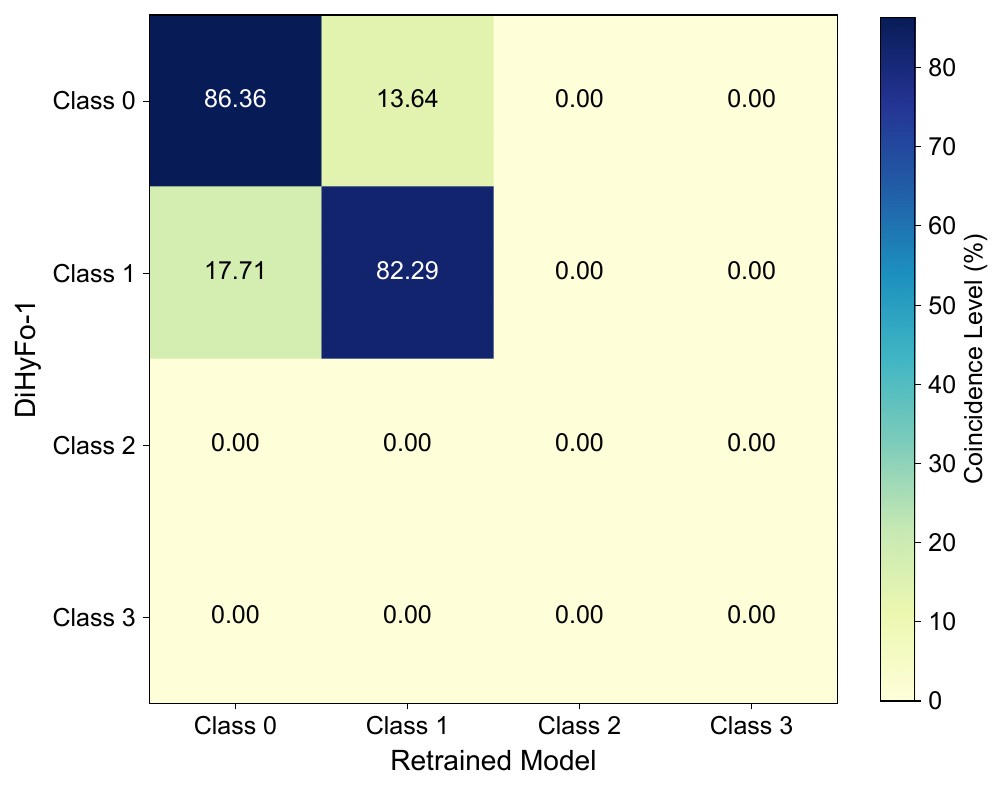}
        \caption{Predictions on test set with $D_f=\{2,3\}$}
        \label{fig:dihyfo1_mnist4_cm_whole_23}
    \end{subfigure}
    \hfill
    % Second subfigure
    \begin{subfigure}[b]{0.49\textwidth}
        \centering
        \includegraphics[width=\textwidth]{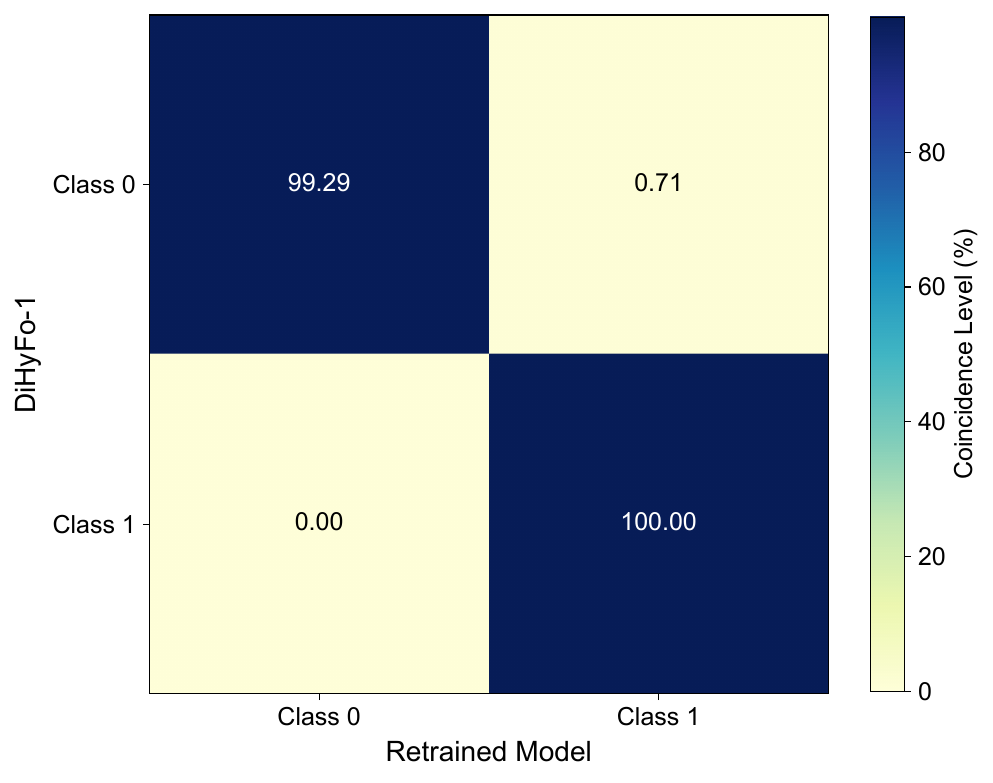}
        \caption{Predictions on $D_r$ with $D_f=\{2,3\}$}
        \label{fig:dihyfo1_mnist4_cm_retain_23}
    \end{subfigure}

    \caption{Comparison of predictions between an unlearned model sampled with DiHyFo-1 and the retrained model on MNIST-4.}
    \label{fig:dihyfo1_mnist4_cm}
\end{figure}

\begin{figure}[H]
    \vspace{1.5in}
    \centering
    % First subfigure
    \begin{subfigure}[b]{0.49\textwidth}
        \centering
        \includegraphics[width=\textwidth]{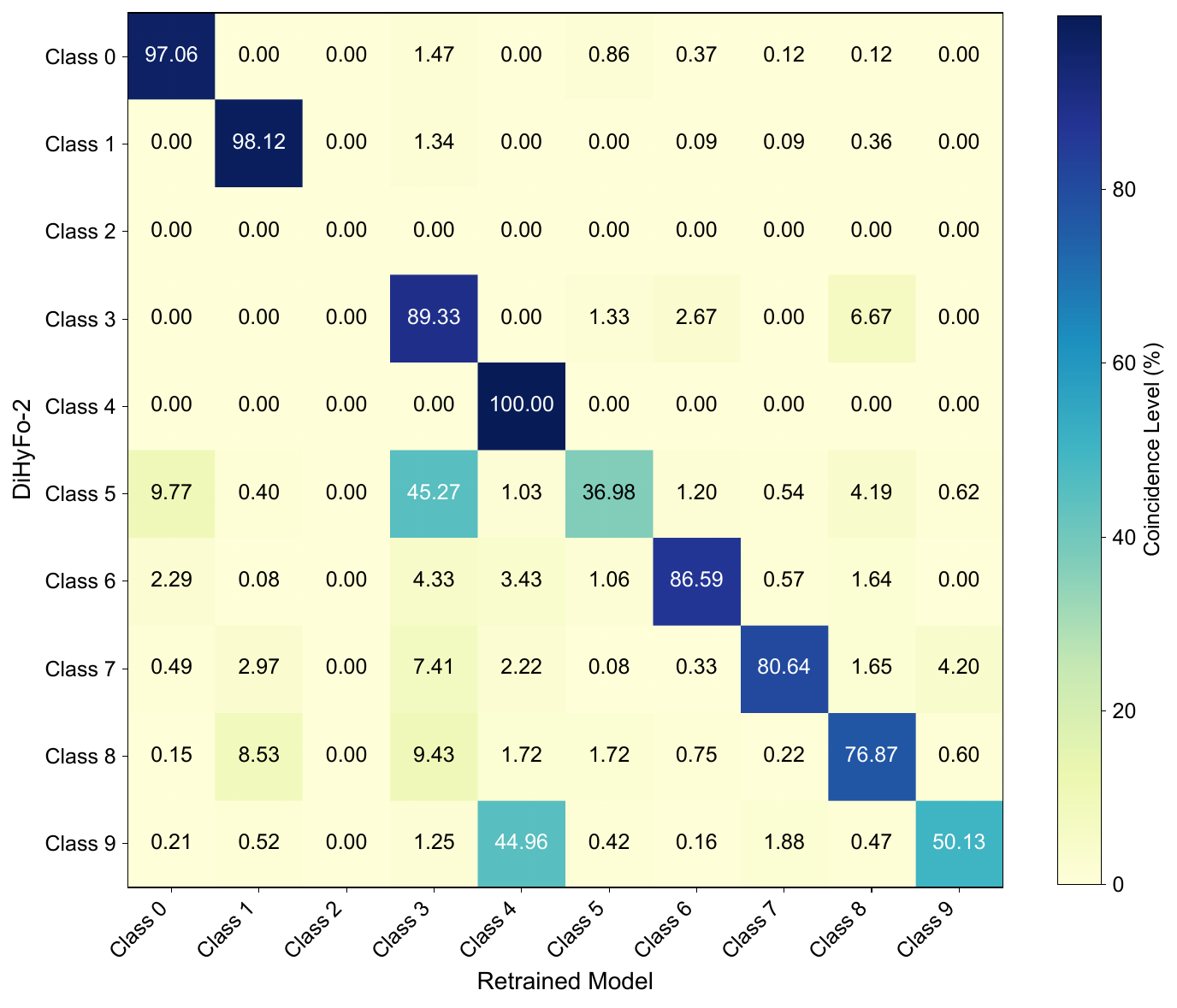}
        \caption{Predictions on test set with $Df=\{2\}$}
        \label{fig:dihyfo2_mnist_cm_whole_2}
    \end{subfigure}
    \hfill
    % Second subfigure
    \begin{subfigure}[b]{0.49\textwidth}
        \centering
        \includegraphics[width=\textwidth]{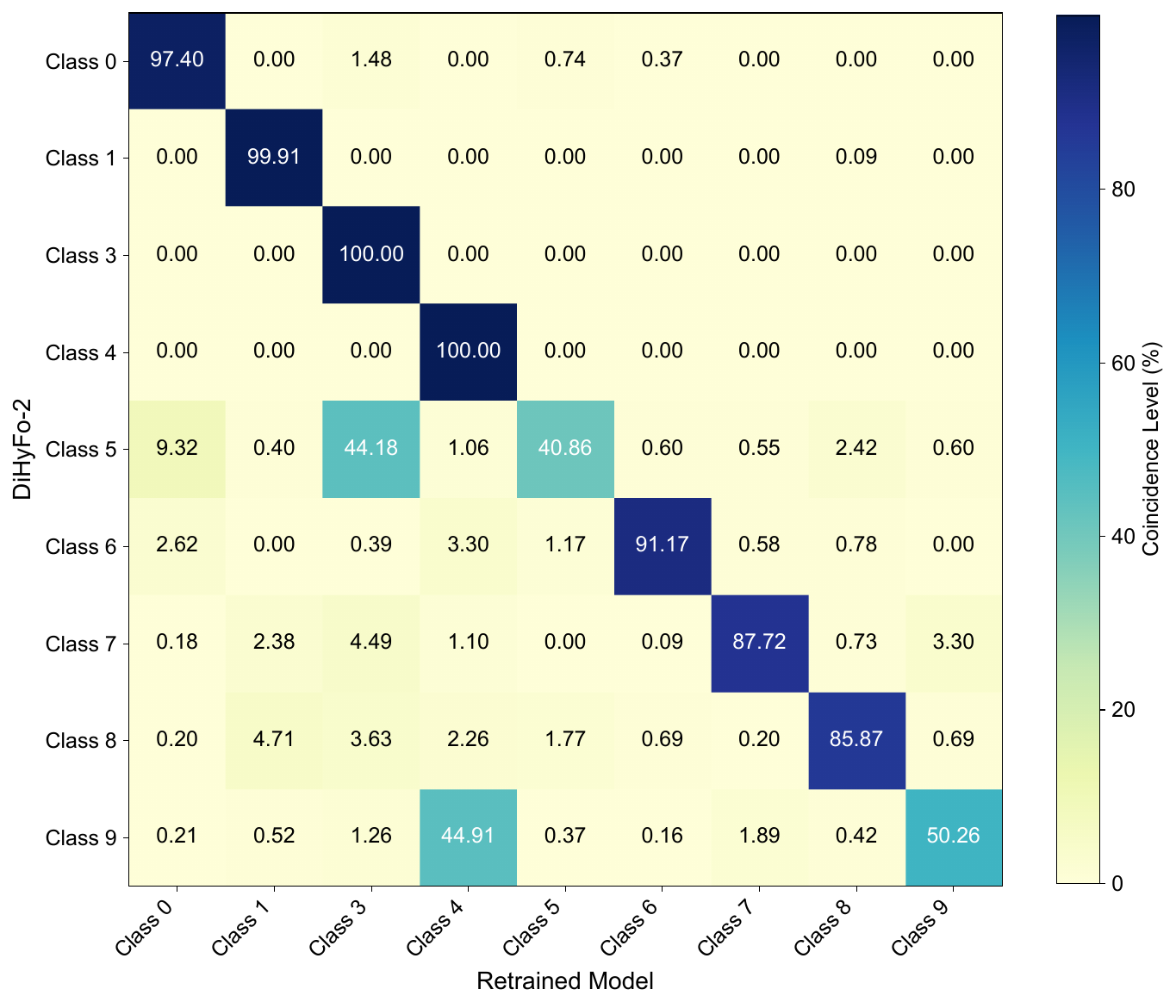}
        \caption{Predictions on $D_r$ with $D_f=\{2\}$}
        \label{fig:dihyfo2_mnist_cm_retain_2}
    \end{subfigure}

    \centering
    % First subfigure
    \begin{subfigure}[b]{0.49\textwidth}
        \centering
        \includegraphics[width=\textwidth]{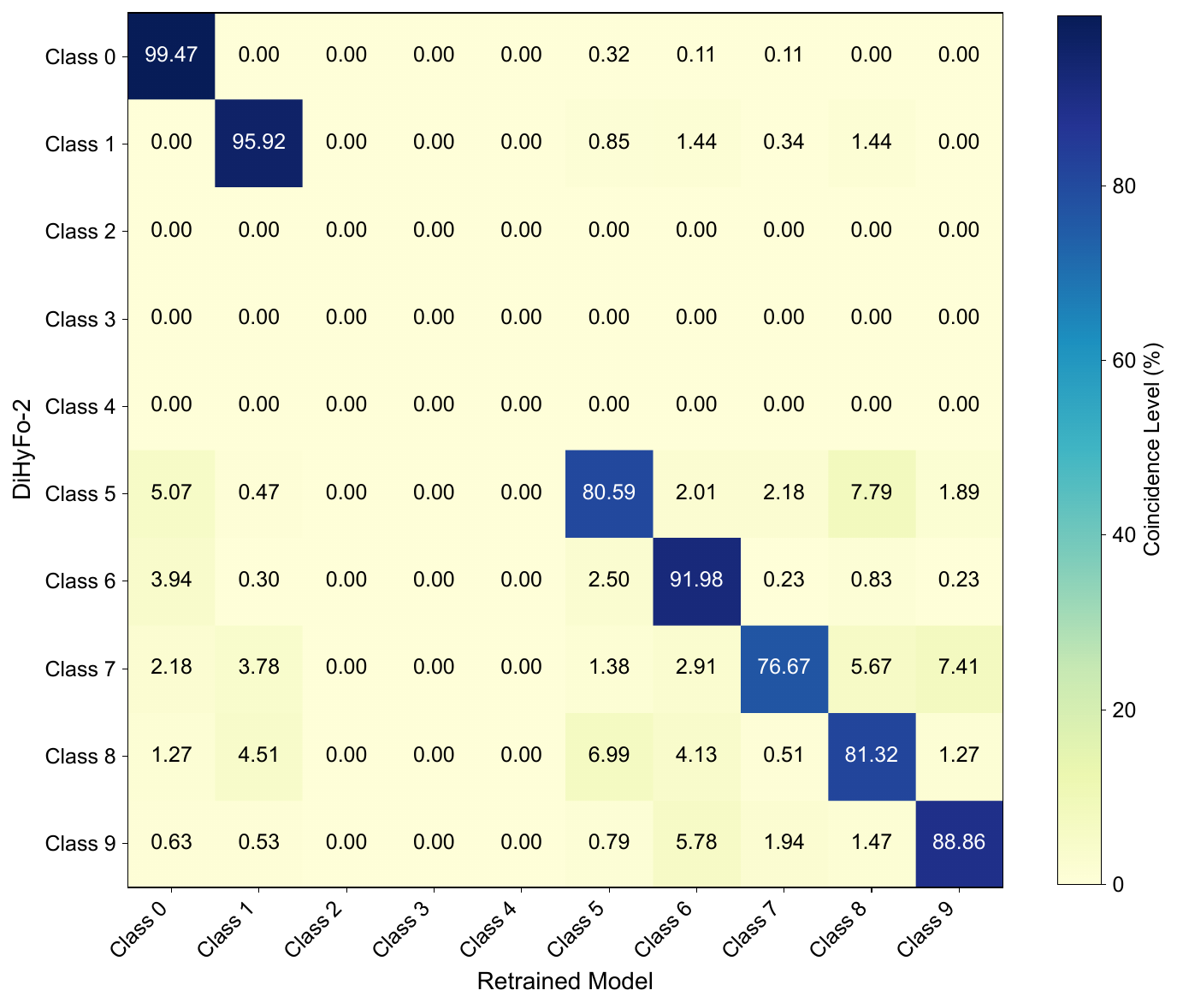}
        \caption{Predictions on test set with $D_f=\{2,3,4\}$}
        \label{fig:dihyfo2_mnist_cm_whole_234}
    \end{subfigure}
    \hfill
    % Second subfigure
    \begin{subfigure}[b]{0.49\textwidth}
        \centering
        \includegraphics[width=\textwidth]{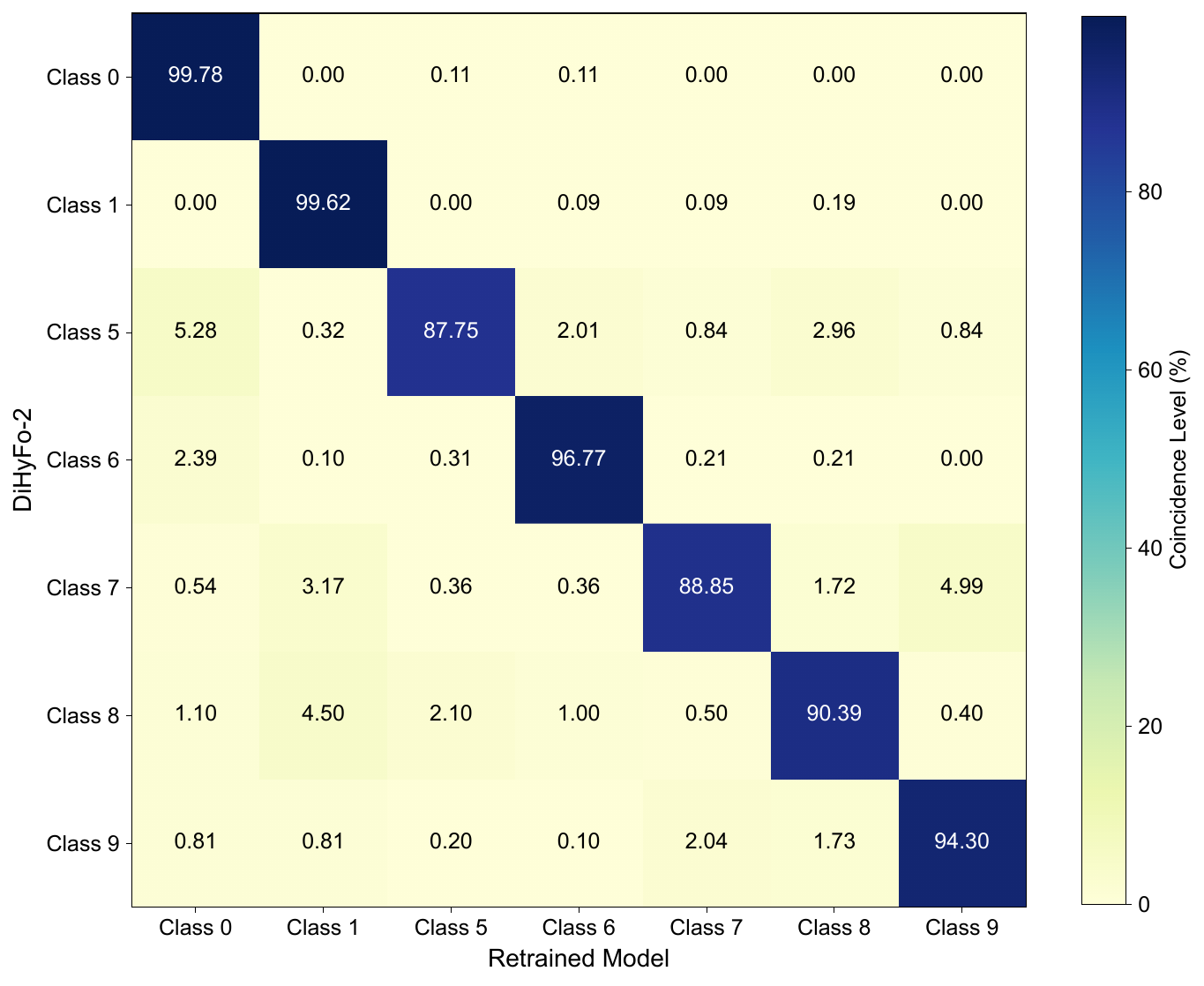}
        \caption{Predictions on $D_r$ with $D_f=\{2,3,4\}$}
        \label{fig:dihyfo2_mnist_cm_retain_234}
    \end{subfigure}

    \caption{Comparison of predictions between an unlearned model sampled with DiHyFo-2 and the retrained model on MNIST.}
    \label{fig:dihyfo2_mnist_cm}
\end{figure}

\begin{figure}[H]
    \vspace{1.5in}
    \centering
    % First subfigure
    \begin{subfigure}[b]{0.49\textwidth}
        \centering
        \includegraphics[width=\textwidth]{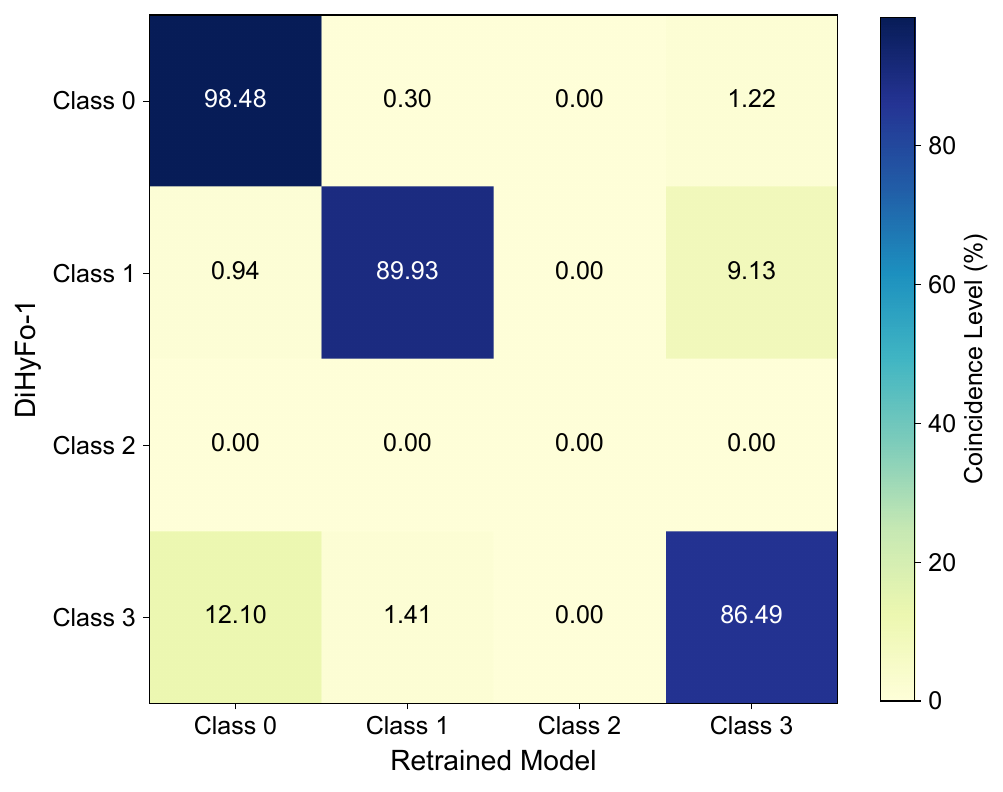}
        \caption{Predictions on test set with $D_f=\{2\}$}
        \label{fig:dihyfo2_mnist4_cm_whole_2}
    \end{subfigure}
    \hfill
    % Second subfigure
    \begin{subfigure}[b]{0.49\textwidth}
        \centering
        \includegraphics[width=\textwidth]{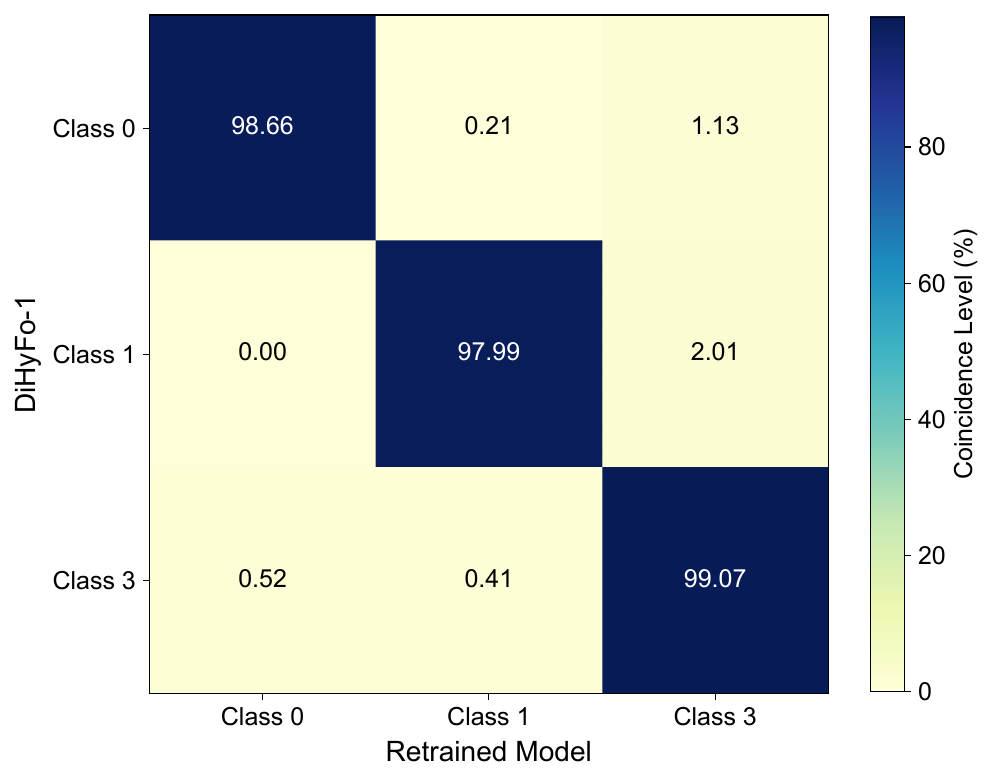}
        \caption{Predictions on $D_r$ with $D_f=\{2\}$}
        \label{fig:dihyfo2_mnist4_cm_retain_2}
    \end{subfigure}

    \centering
    % First subfigure
    \begin{subfigure}[b]{0.49\textwidth}
        \centering
        \includegraphics[width=\textwidth]{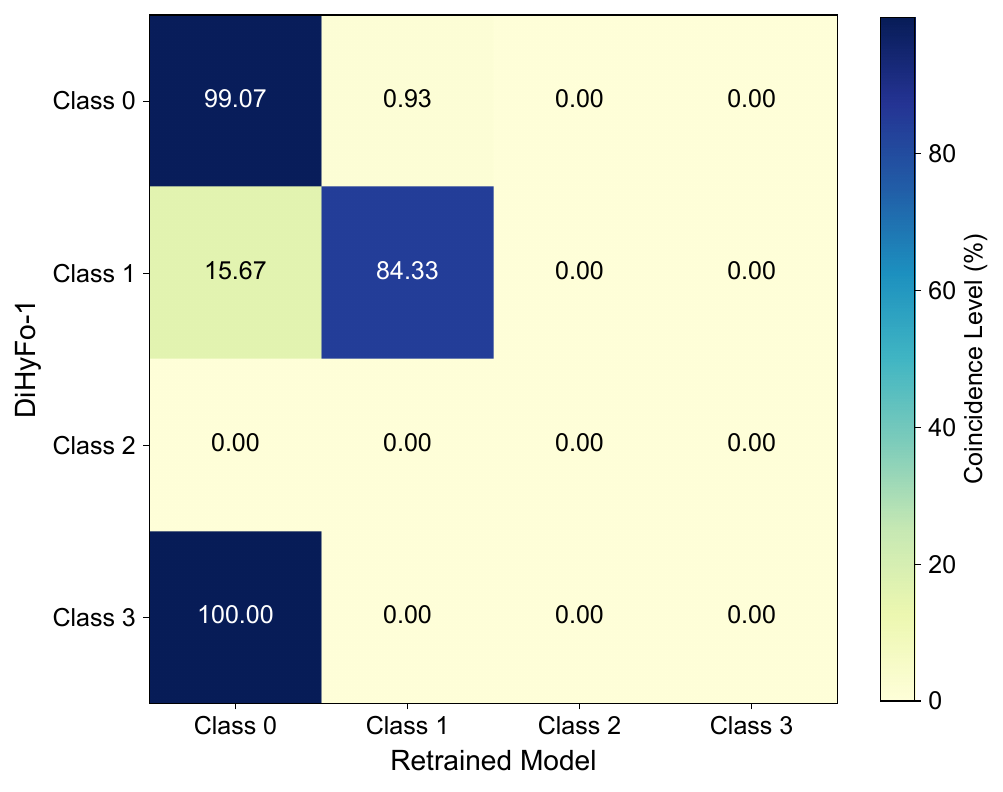}
        \caption{Predictions on test set with $D_f=\{2,3\}$}
        \label{fig:dihyfo2_mnist4_cm_whole_23}
    \end{subfigure}
    \hfill
    % Second subfigure
    \begin{subfigure}[b]{0.49\textwidth}
        \centering
        \includegraphics[width=\textwidth]{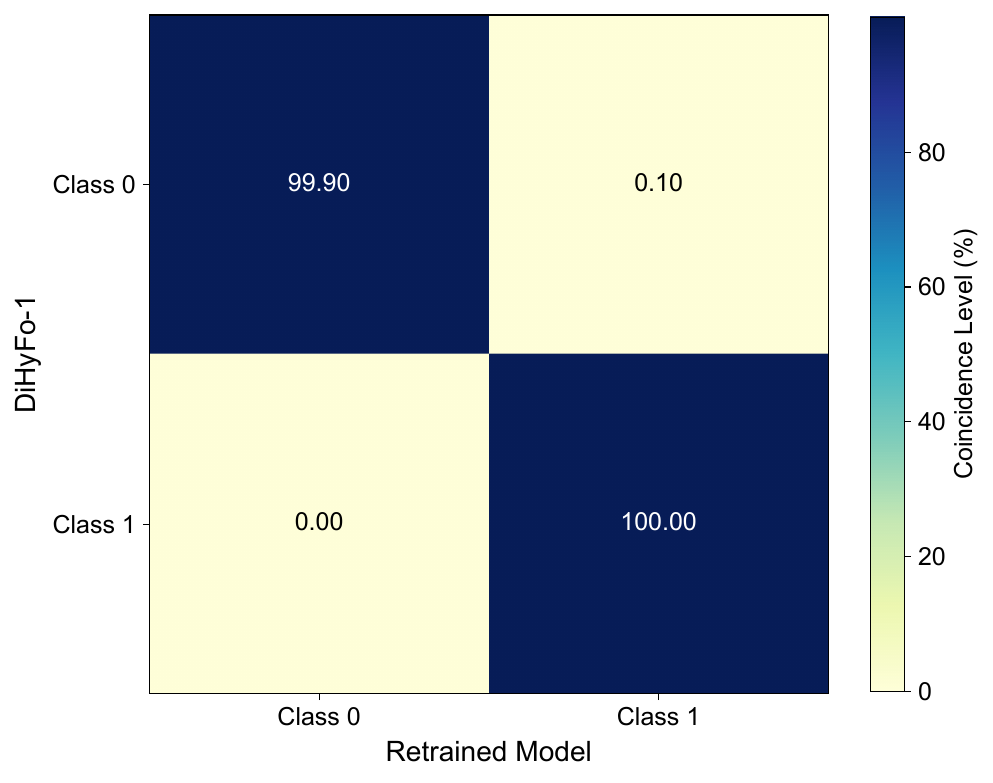}
        \caption{Predictions on $D_r$ with $D_f=\{2,3\}$}
        \label{fig:dihyfo2_mnist4_cm_retain_23}
    \end{subfigure}

    \caption{Comparison of predictions between an unlearned model sampled with DiHyFo-2 and the retrained model on MNIST-4.}
    \label{fig:dihyfo2_mnist4_cm}
\end{figure}

Taking as reference the predictions of the retrained model, we also compare the output spaces using the confusion matrix between the sampled models and the corresponding retrained model on the retain and complete test set. Figures \ref{fig:dihyfo1_mnist_cm}, \ref{fig:dihyfo1_mnist4_cm}, \ref{fig:dihyfo2_mnist_cm}, and \ref{fig:dihyfo2_mnist4_cm} present the results for DiHyFo-1 and DiHyFo-2 on MNIST-4 and MNIST. The forget set is not included in this comparison as it mainly contains matrices with zeros across all entries. This serves as a visual confirmation of commented findings using the evaluation metrics, obtaining zeroes in the forget set classes.

Overall, the evaluations suggest that the presented DiHyFo models accomplish some relevant desired conditions for an unlearning algorithm \citep{nguyen2022survey, zhang2023review,ginart2019making,chundawat2023zero,tarun2023fast,guo2019certified,yoon2022few,micaelli2019zero,golatkar2020eternal,
he2021deepobliviate, Bourtoule2021MachineUnlearning, shuang2024forgetting}. They 
were consistent, timely at sampling unlearned models, provided guarantees and verification of unlearning, but were model intrinsic, with potential recoverability of knowledge, and had issues with scalability.

\begin{table}[H]
\centering
\begin{minipage}[b]{\textwidth}
\caption{Individual unlearned models performance on MNIST-4 under diverse forgetting scenarios. Average over 5 sampled unlearned models.}
\vspace{1em}
\centering
\label{table:inv_metrics_mnist4}
\adjustbox{width=0.75\textwidth}{
\begin{tabular}{l c c c c c c}
\toprule
\multirow{2}{*}{MNIST-4} & \multicolumn{2}{c}{Sampled with DiHyfo-1} & \multicolumn{2}{c}{Sampled with DiHyfo-2} & \multicolumn{2}{c}{Retrained Model} \\ 
\cmidrule(lr){2-3} \cmidrule(lr){4-5} \cmidrule(lr){6-7}
& $D_f$=\{2\} & $D_f$=\{2,3\} & $D_f$=\{2\} & $D_f$=\{2,3\} & $D_f$=\{2\} & $D_f$=\{2,3\} \\
\midrule
Accuracy on $D_r$ \((\uparrow)\) & 0.9837 & 0.9983 & 0.9834 & 0.9995 & 0.9938 & 0.9991 \\
Accuracy on $D_f$ \((\downarrow)\) & 0.0000 & 0.0000 & 0.0000 & 0.0000 & 0.0000 & 0.0000 \\
MIA \((\downarrow)\) & 0.6260 & 0.6484 & 0.4457 & 0.3827 & 0.4171 & 0.3398 \\
\bottomrule
\end{tabular}
}
\end{minipage}%
\end{table}

\begin{table}[H]
\centering
\begin{minipage}[b]{\textwidth}
\caption{Assessment of output alignment and behavioral consistency between the indicated sampled unlearned model and its corresponding retrained model on MNIST-4 under diverse forgetting scenarios. Average over 5 sampled unlearned models.}
\vspace{1em}
\centering
\label{table:comp_metrics_mnist4}
\adjustbox{width=0.7\textwidth}{%
\begin{tabular}{l c c c c}
\toprule
\multirow{2}{*}{MNIST-4} & \multicolumn{2}{c}{Sampled with DiHyfo-1} & \multicolumn{2}{c}{Sampled with DiHyfo-2} \\ 
\cmidrule(lr){2-3} \cmidrule(lr){4-5}
& $D_f$=\{2\} & $D_f$=\{2,3\} & $D_f$=\{2\} & $D_f$=\{2,3\} \\
\midrule
Outputs Overlap on $D_r$ \((\uparrow)\) & 0.9851 & 0.9983 & 0.9848 & 0.9995 \\
Outputs Overlap on $D_f$ \((-)\) & 0.3473 & 0.7106 & 0.6800 & 0.7751 \\
Outputs Overlap on Test Set \((\uparrow)\) & 0.8263 & 0.8562 & 0.9094 & 0.8898 \\
Unlearning Score ($\varphi$) \((\uparrow)\) & 0.8550 & 0.9327 & 0.6874 & 0.9607 \\
\bottomrule
\end{tabular}
}
\end{minipage}%
\end{table}

\newpage

\subsection{Additional training results}
\label{subsec:training}

\begin{figure}[H]
%\vspace{-10em}
    \centering
    % First subfigure
    \begin{subfigure}[b]{0.49\textwidth}
        \centering
        \includegraphics[width=\textwidth]{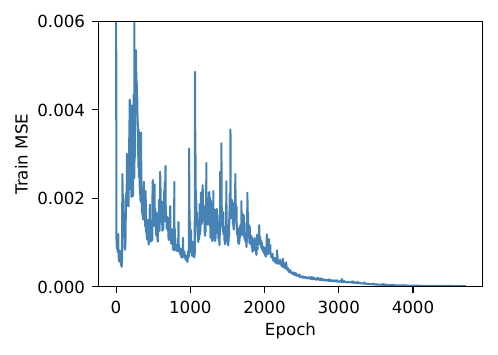}
        \caption{DiHyFo-1, train}
        \label{fig:dihyfo1_mnist4_train_loss}
    \end{subfigure}
    \hfill
    % Second subfigure
    \begin{subfigure}[b]{0.49\textwidth}
        \centering
        \includegraphics[width=\textwidth]{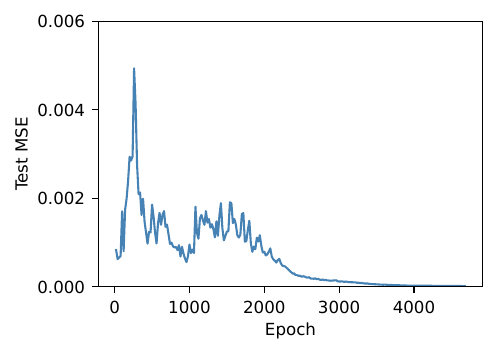}
        \caption{DiHyFo-1, test}
        \label{fig:dihyfo1_mnist4_test_loss}
    \end{subfigure}

    \hfill
    % First subfigure
    \begin{subfigure}[b]{0.49\textwidth}
        \centering
        \includegraphics[width=\textwidth]{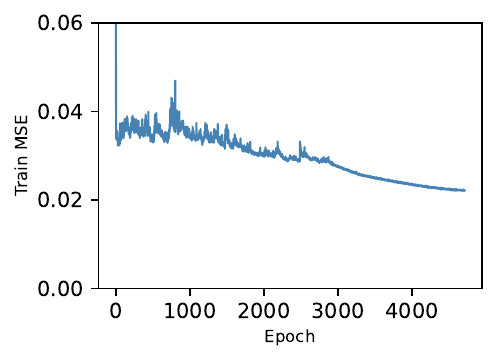}
        \caption{DiHyFo-2, train}
        \label{fig:dihyfo2_mnist4_train_loss}
    \end{subfigure}
    \hfill
    % Second subfigure
    \begin{subfigure}[b]{0.49\textwidth}
        \centering
        \includegraphics[width=\textwidth]{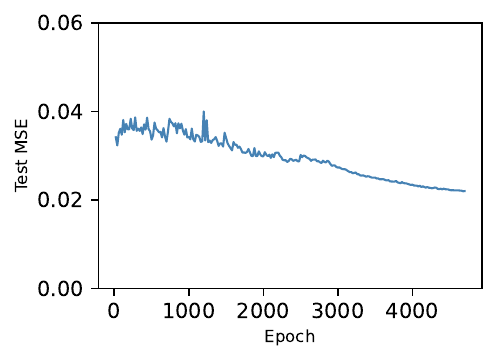}
        \caption{DiHyFo-2, test}
        \label{fig:dihyfo2_mnist4_test_loss}
    \end{subfigure}

    \caption{DiHyFo models learning curves for MNIST-4.}
    \label{fig:dihyfo2_mnist4_lc}
\end{figure}

\begin{figure}[htbp]
    \centering
    % First subfigure
    \begin{subfigure}[b]{0.48\textwidth}
        \centering
        \includegraphics[width=\textwidth]{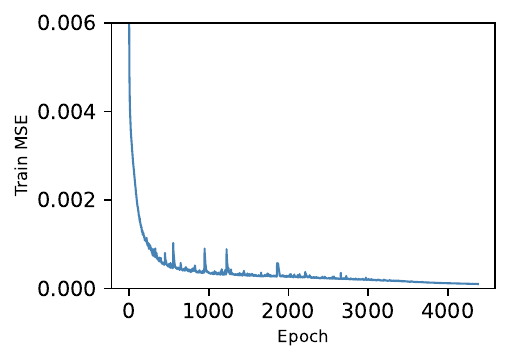}
        \caption{DiHyFo-1, train}
        \label{fig:dihyfo1_mnist_train_loss}
    \end{subfigure}
    \hfill
    % Second subfigure
    \begin{subfigure}[b]{0.48\textwidth}
        \centering
        \includegraphics[width=\textwidth]{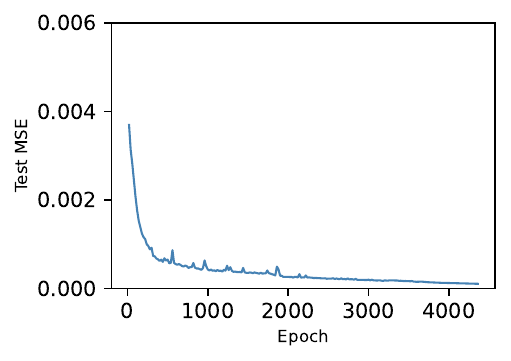}
        \caption{DiHyFo-1, test}
        \label{fig:dihyfo1_mnist_test_loss}
    \end{subfigure}

    \hfill
    % First subfigure
    \begin{subfigure}[b]{0.48\textwidth}
        \centering
        \includegraphics[width=\textwidth]{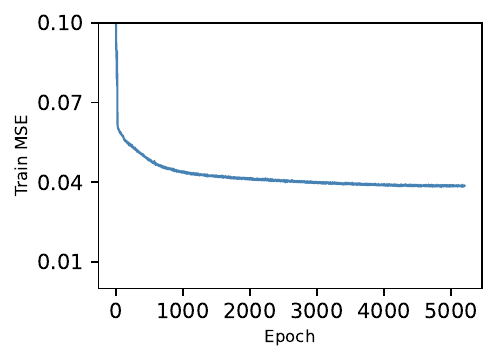}
        \caption{DiHyFo-2, train}
        \label{fig:dihyfo2_mnist_train_loss}
    \end{subfigure}
    \hfill
    % Second subfigure
    \begin{subfigure}[b]{0.48\textwidth}
        \centering
        \includegraphics[width=\textwidth]{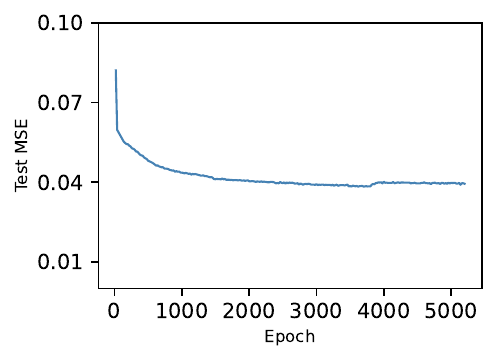}
        \caption{DiHyFo-2, test}
        \label{fig:dihyfo2_mnist_test_loss}
    \end{subfigure}

    \caption{DiHyFo models learning curves for MNIST.}
    \label{fig:dihyfo2_mnist_lc}
\end{figure}

Learning curves for both models decrease until convergence and show a similar behavior during training and testing, Figures \ref{fig:dihyfo2_mnist4_lc} and \ref{fig:dihyfo2_mnist_lc}.

Most experiments presented positive prompt alignment scores, indicating that the generated parameters obtain class losses that closely align with the desired class losses. However, there are cases in which even when the prompt alignment increases and stabilizes it does not achieve to surpass zero, contrasting with what is shown in the correlation and direct comparison plots. As illustrated in Table \ref{table:prompt_alignment_correlation}, in several cases this behavior is due to some generated models having high negative scores that pull the average down, suggesting that the performance of the models as a group is quite varied, with most models performing good but some models performing quite poorly in terms of generating parameters for the desired losses.

On the other hand, most models show high correlation between observed and desired losses, even for models with a low prompt alignment. This suggests that while the models may not always match the exact value of the losses, they generally align with the appropriate loss direction.

Results in Table \ref{table:prompt_alignment_correlation} were computed using the sampled models with DiHyFo-2 on MNIST-4 during the last epoch, and can be contrasted with the corresponding direct comparison plot. The general loss direction alignment does not always translate into accurate loss matching. Considering the extreme negative values as outliers or setting a lower bound of 0 on the prompt alignment makes it resemble more closely the correlation results. Indeed, the models are capturing the direction of the relationship well but might struggle with the precision of their predictions.

\begin{table}[H]
\centering
\begin{minipage}[b]{\textwidth}
\caption{Prompt alignment and correlation of losses for class 2 of MNIST-4 obtained by models sampled using DiHyFo-2.}
%\vspace{1em}
\centering
\label{table:prompt_alignment_correlation}
\adjustbox{width=0.68\textwidth}{
\scriptsize % Reduce font size
\begin{tabular}{l c c}
\toprule
\textbf{Model Number} & \textbf{Prompt Alignment} & \textbf{Correlation (Pearson)} \\ 
\midrule
0  & -6.4499 & 0.4931 \\ 
1  &  0.9443 & 0.9780 \\ 
2  & -6.8617 & 0.5607 \\ 
3  &  0.8824 & 0.9448 \\ 
4  & -0.7479 & 0.8891 \\ 
5  &  0.9587 & 0.9819 \\ 
6  &  0.7846 & 0.9277 \\ 
7  &  0.8943 & 0.9486 \\ 
8  &  0.8392 & 0.9457 \\ 
9  &  0.8056 & 0.9357 \\ 
10 &  0.5366 & 0.8465 \\ 
11 &  0.8466 & 0.9617 \\ 
12 &  0.8935 & 0.9621 \\ 
13 &  0.7153 & 0.9052 \\ 
14 &  0.5695 & 0.8738 \\ 
15 &  0.6855 & 0.8934 \\ 
16 &  0.8757 & 0.9361 \\ 
17 & -1.8833 & -0.2664 \\ 
18 &  0.5878 & 0.9455 \\ 
19 & -10.7378 & -0.1768 \\ 
20 &  0.9206 & 0.9646 \\ 
21 &  0.7484 & 0.8999 \\ 
22 &  0.4875 & 0.8033 \\ 
23 &  0.9187 & 0.9683 \\ 
\midrule
Average & -0.4911 & 0.7968 \\ 
\bottomrule
\end{tabular}
}
\end{minipage}%
\end{table}

\begin{figure}[H]
\vspace{-0.15in}
    \centering
    % First subfigure
    \begin{subfigure}[b]{0.48\textwidth}
        \centering
        \includegraphics[width=\textwidth]{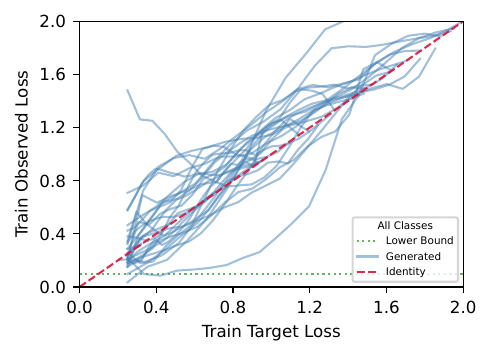}
        \caption[dihyfo1_mnist4_obs_train2]{Class 2 - train}
        \label{fig:dihyfo1_mnist4_obs_train2}
    \end{subfigure}
    \hfill
    % Second subfigure
    \begin{subfigure}[b]{0.48\textwidth}
        \centering
        \includegraphics[width=\textwidth]{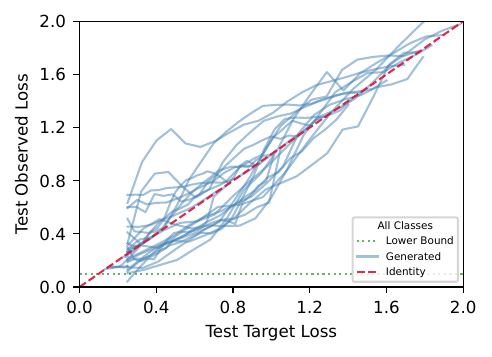}
        \caption[dihyfo1_mnist4_obs_test2]{Class 2 - test}
        \label{fig:dihyfo1_mnist4_obs_test2}
    \end{subfigure}
    \hfill
    \centering
    % First subfigure
    \begin{subfigure}[b]{0.48\textwidth}
        \centering
        \includegraphics[width=\textwidth]{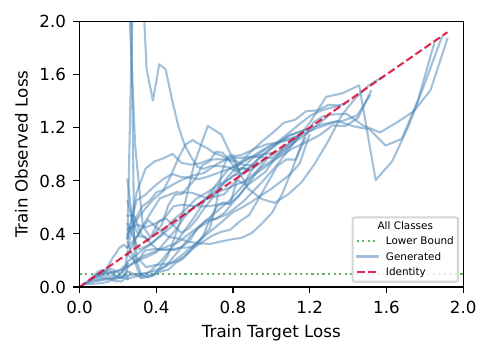}
        \caption[dihyfo1_mnist4_obs_train0]{Class 0 (pivot) - train}
        \label{fig:dihyfo1_mnist4_obs_train0}
    \end{subfigure}
    \hfill
    % Second subfigure
    \begin{subfigure}[b]{0.48\textwidth}
        \centering
        \includegraphics[width=\textwidth]{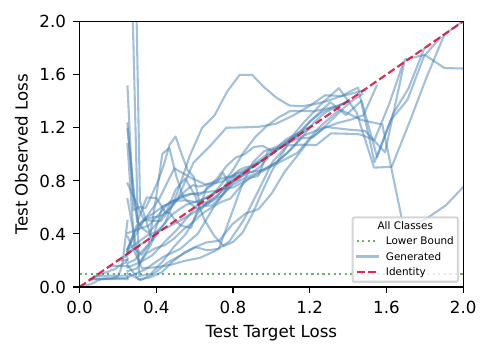}
        \caption[dihyfo1_mnist4_obs_test0]{Class 0 (pivot) - test}
        \label{fig:dihyfo1_mnist4_obs_test0}
    \end{subfigure}
    
    \caption{Comparison of losses obtained by the parameters generated with DiHyFo-1 for MNIST-4.}
    \label{fig:obs_des_dihyfo1_mnist4}
\end{figure}

\begin{figure}[H]
\vspace{-0.3in}
    \centering
    % First subfigure
    \begin{subfigure}[b]{0.462\textwidth}
        \centering
        \includegraphics[width=\textwidth]{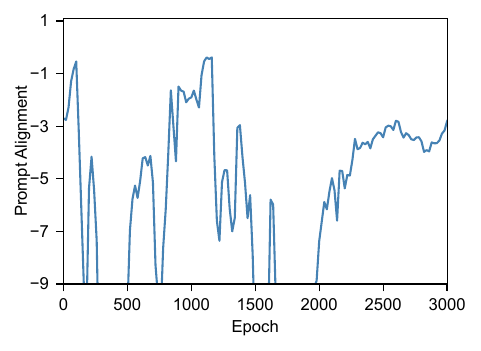}
        \caption[DiHyFo-1 Prompt Alignment on Class 2]{Class 2}
        \label{fig:dihyfo1_mnist4_c2r2}
    \end{subfigure}
    \hfill
    % Second subfigure
    \begin{subfigure}[b]{0.47\textwidth}
        \centering
        \includegraphics[width=\textwidth]{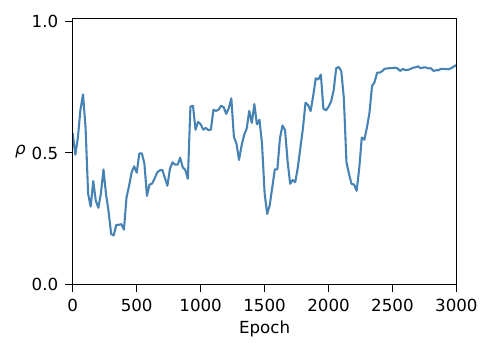}
        \caption[DiHyFo-1 Correlation on Class 2]{Class 2}
        \label{fig:dihyfo1_mnist4_c2pearson_corr}
    \end{subfigure}
    \hfill
    
    % First subfigure
    \begin{subfigure}[b]{0.462\textwidth}
        \centering
        \includegraphics[width=\textwidth]{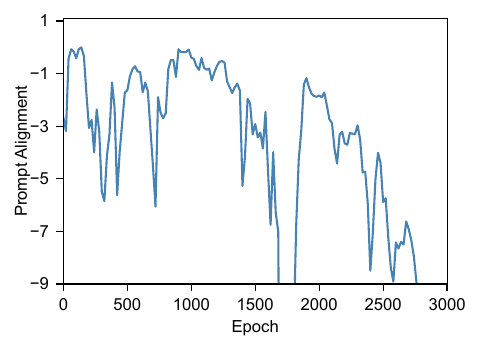}
        \caption[DiHyFo-1 Prompt Alignment on Class 0]{Class 0 (pivot)}
        \label{fig:dihyfo1_mnist4_c0r2}
    \end{subfigure}
    \hfill
    % Second subfigure
    \begin{subfigure}[b]{0.47\textwidth}
        \centering
        \includegraphics[width=\textwidth]{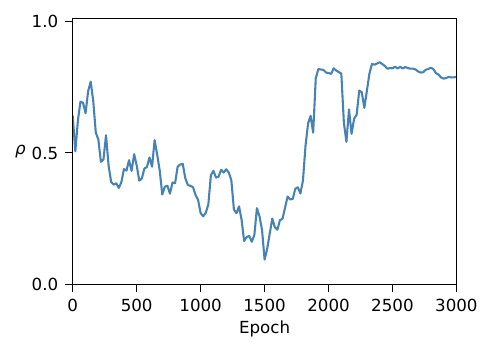}
        \caption[DiHyFo-1 Correlation on Class 0]{Class 0 (pivot)}
        \label{fig:dihyfo1_mnist4_c0pearson_corr}
    \end{subfigure}
    
    \caption{Prompt alignment and correlation of the losses obtained by the parameters generated with DiHyFo-1 for MNIST-4.}
    \label{fig:pa_corr_dihyfo1_mnist4}
\end{figure}

\begin{figure}[H]
\vspace{-0.15in}
    \centering
    % First subfigure
    \begin{subfigure}[b]{0.48\textwidth}
        \centering
        \includegraphics[width=\textwidth]{figs/appen/dihyfo2_mnist4_obs_train2.pdf}
        \caption{Losses Comparison for class 2 - train}
        \label{fig:dihyfo2_mnist4_loss_train2}
    \end{subfigure}
    \hfill
    % Second subfigure
    \begin{subfigure}[b]{0.48\textwidth}
        \centering
        \includegraphics[width=\textwidth]{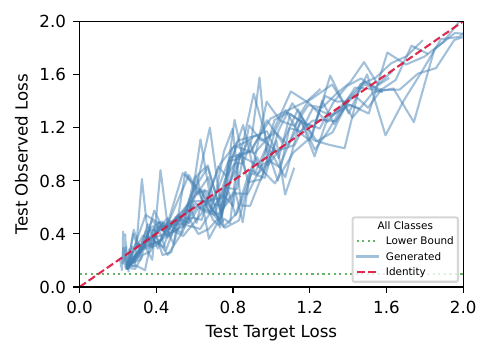}
        \caption{Losses comparison for class 2 - test}
        \label{fig:dihyfo2_mnist4_obs_test2}
    \end{subfigure}
    \hfill
    \centering
    % First subfigure
    \begin{subfigure}[b]{0.48\textwidth}
        \centering
        \includegraphics[width=\textwidth]{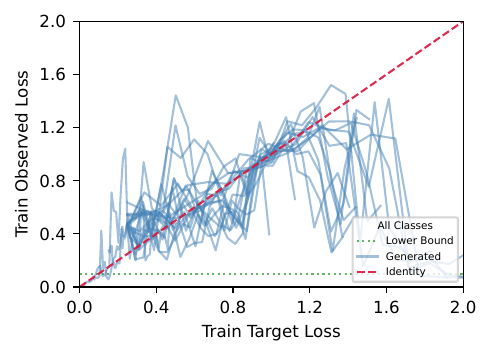}
        \caption{Losses comparison for class 1 (pivot) - train}
        \label{fig:dihyfo2_mnist4_obs_train0}
    \end{subfigure}
    \hfill
    % Second subfigure
    \begin{subfigure}[b]{0.48\textwidth}
        \centering
        \includegraphics[width=\textwidth]{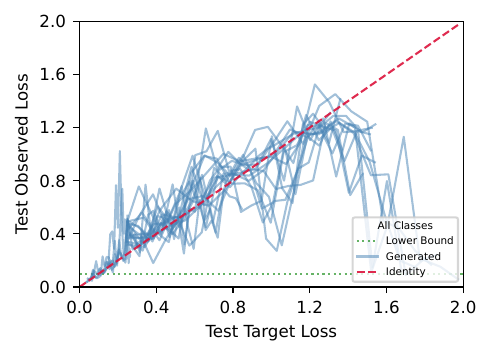}
        \caption{Losses comparison for class 1 (pivot) - test}
        \label{fig:dihyfo2_mnist4_obs_test1}
    \end{subfigure}
    
    \caption{Comparison of losses obtained by the parameters generated with DiHyFo-2 for MNIST-4.}
    \label{fig:obs_des_dihyfo2_mnist4}
\end{figure}

\begin{figure}[H]
\vspace{-0.3in}
    \centering
    % First subfigure
    \begin{subfigure}[b]{0.462\textwidth}
        \centering
        \includegraphics[width=\textwidth]{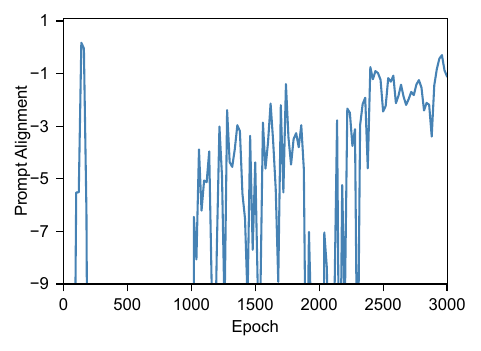}
        \caption{Prompt alignment for class 2}
        \label{fig:dihyfo2_mnist4_c2r2}
    \end{subfigure}
    \hfill
    % Second subfigure
    \begin{subfigure}[b]{0.47\textwidth}
        \centering
        \includegraphics[width=\textwidth]{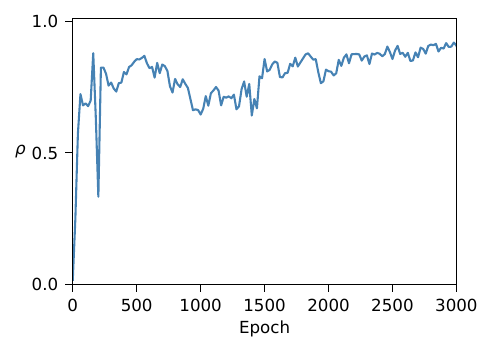}
        \caption{Correlation for class 2}
        \label{fig:dihyfo2_mnist4_c2pearson_corr}
    \end{subfigure}
    \hfill
    \centering
    % First subfigure
    \begin{subfigure}[b]{0.462\textwidth}
        \centering
        \includegraphics[width=\textwidth]{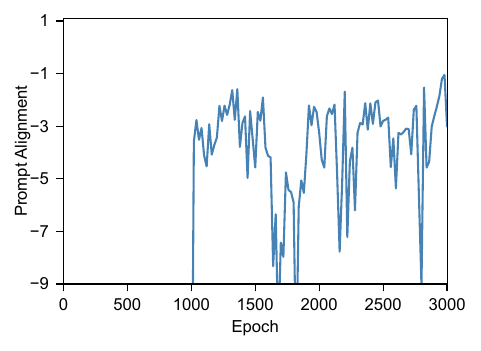}
        \caption{Prompt alignment for class 1 (pivot)}
        \label{fig:dihyfo2_mnist4_c1r2}
    \end{subfigure}
    \hfill
    % Second subfigure
    \begin{subfigure}[b]{0.47\textwidth}
        \centering
        \includegraphics[width=\textwidth]{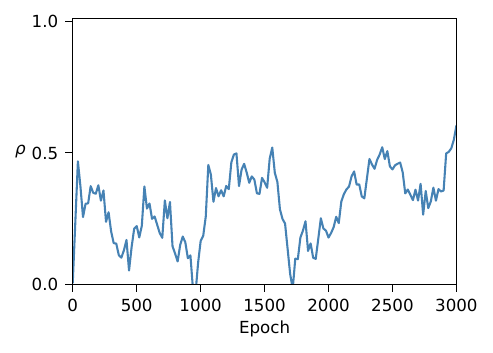}
        \caption{Correlation for class 1 (pivot)}
        \label{fig:dihyfo2_mnist4_c0pearson_corr}
    \end{subfigure}
    
    \caption{Prompt alignment and correlation of losses obtained by the parameters generated with DiHyFo-2 for MNIST-4.}
    \label{fig:pa_corr_dihyfo2_mnist4}
\end{figure}

\begin{figure}[H]
\vspace{1in}
    \centering
    % First subfigure
    \begin{subfigure}[b]{0.48\textwidth}
        \centering
        \includegraphics[width=\textwidth]{figs/appen/dihyfo1_mnist_obs_train2.pdf}
        \caption{Losses comparison for class 2 - train}
        \label{fig:dihyfo1_mnist_obs_train2}
    \end{subfigure}
    \hfill
    % Second subfigure
    \begin{subfigure}[b]{0.48\textwidth}
        \centering
        \includegraphics[width=\textwidth]{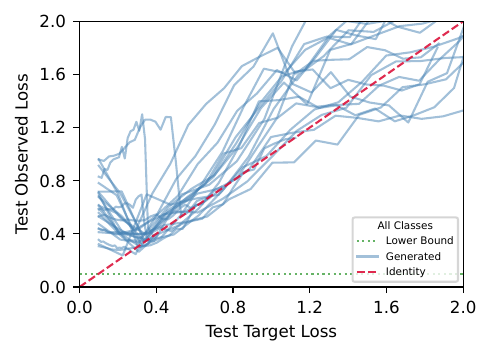}
        \caption{Losses comparison for class 2 - test}
        \label{fig:dihyfo1_mnist_obs_test2}
    \end{subfigure}
    \hfill

    \centering
    % First subfigure
    \begin{subfigure}[b]{0.48\textwidth}
        \centering
        \includegraphics[width=\textwidth]{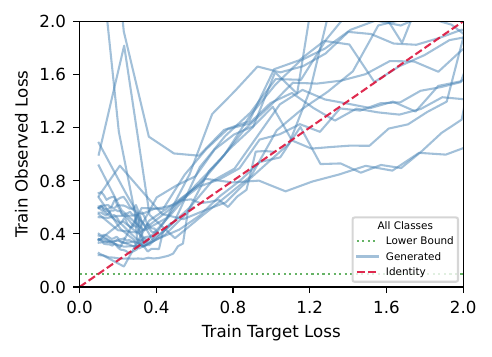}
        \caption{Losses comparison for class 3 - train}
        \label{fig:dihyfo1_mnist_obs_train3}
    \end{subfigure}
    \hfill
    % Second subfigure
    \begin{subfigure}[b]{0.48\textwidth}
        \centering
        \includegraphics[width=\textwidth]{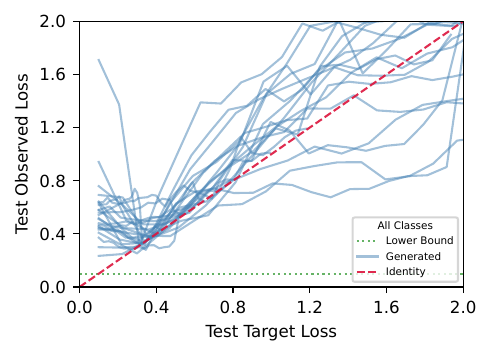}
        \caption{Losses comparison for class 3 - test}
        \label{fig:dihyfo1_mnist_obs_test3}
    \end{subfigure}
    \hfill
    
    \centering
    % First subfigure
    \begin{subfigure}[b]{0.48\textwidth}
        \centering
        \includegraphics[width=\textwidth]{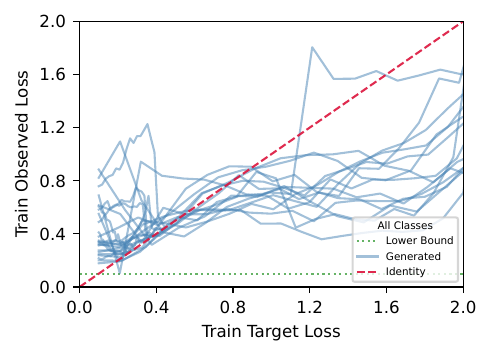}
        \caption{Losses comparison for class 9 (pivot) - train}
        \label{fig:dihyfo1_mnist_obs_train9}
    \end{subfigure}
    \hfill
    % Second subfigure
    \begin{subfigure}[b]{0.48\textwidth}
        \centering
        \includegraphics[width=\textwidth]{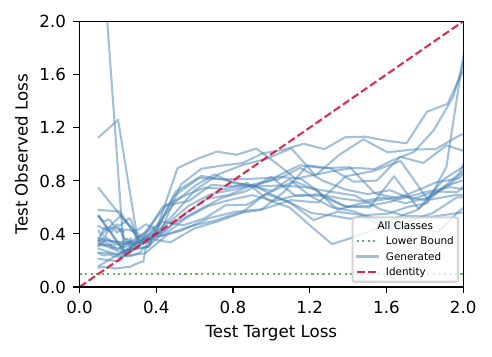}
        \caption{Losses Comparison for class 9 (pivot) - test}
        \label{fig:dihyfo1_mnist_obs_test9}
    \end{subfigure}

    \caption{Comparison of losses obtained by the parameters generated with DiHyFo-1 for MNIST.}
    \label{fig:obs_des_dihyfo1_mnist}
\end{figure}

\begin{figure}[H]
\vspace{1in}
    \centering
    % First subfigure
    \begin{subfigure}[b]{0.472\textwidth}
        \centering
        \includegraphics[width=\textwidth]{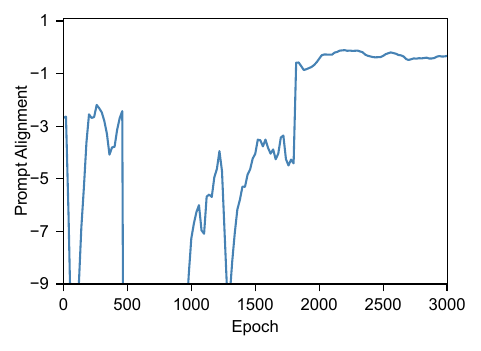}
        \caption{Prompt alignment on class 2}
        \label{fig:dihyfo1_mnist_c2r2}
    \end{subfigure}
    \hfill
    % Second subfigure
    \begin{subfigure}[b]{0.48\textwidth}
        \centering
        \includegraphics[width=\textwidth]{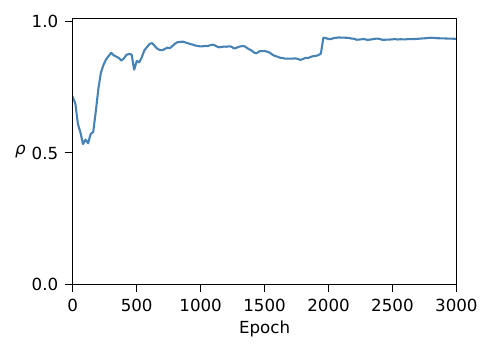}
        \caption{Correlation on class 2}
        \label{fig:dihyfo1_mnist_c2pearson_corr}
    \end{subfigure}
    \hfill

    \centering
    % First subfigure
    \begin{subfigure}[b]{0.472\textwidth}
        \centering
        \includegraphics[width=\textwidth]{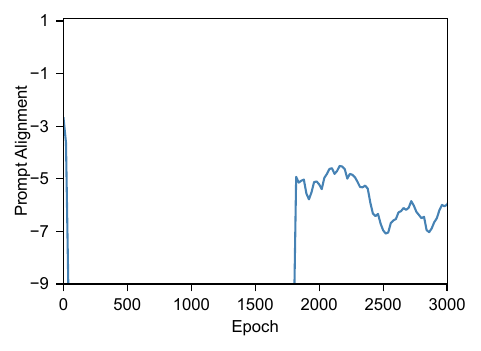}
        \caption{Prompt alignment on class 3}
        \label{fig:dihyfo1_mnist_c3r2}
    \end{subfigure}
    \hfill
    % Second subfigure
    \begin{subfigure}[b]{0.48\textwidth}
        \centering
        \includegraphics[width=\textwidth]{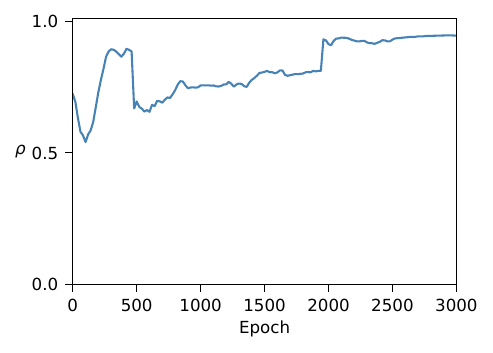}
        \caption{Correlation on class 3}
        \label{fig:dihyfo1_mnist_c3pearson_corr}
    \end{subfigure}
    \hfill
    
    \centering
    % First subfigure
    \begin{subfigure}[b]{0.472\textwidth}
        \centering
        \includegraphics[width=\textwidth]{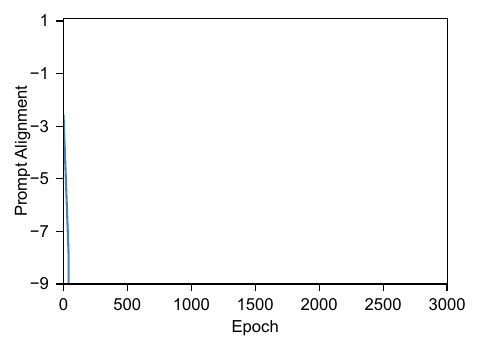}
        \caption{Prompt alignment on class 8 (pivot)}
        \label{fig:dihyfo1_mnist_c8r2}
    \end{subfigure}
    \hfill
    % Second subfigure
    \begin{subfigure}[b]{0.48\textwidth}
        \centering
        \includegraphics[width=\textwidth]{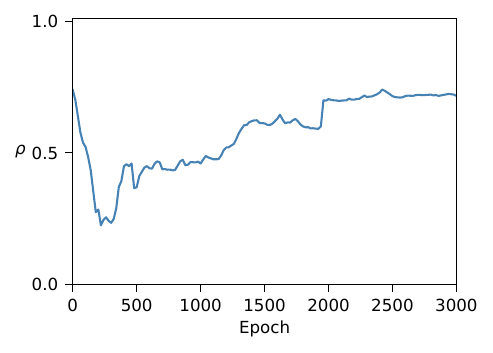}
        \caption{Correlation on class 8 (pivot)}
        \label{fig:dihyfo1_mnist_c8pearson_corr}
    \end{subfigure}
        
    \caption{Prompt alignment and correlation of losses obtained by the parameters generated with DiHyFo-1 for MNIST.}
    \label{fig:pa_corr_dihyfo1_mnist}
\end{figure}

\begin{figure}[H]
\vspace{1in}
    \centering
    % First subfigure
    \begin{subfigure}[b]{0.48\textwidth}
        \centering
        \includegraphics[width=\textwidth]{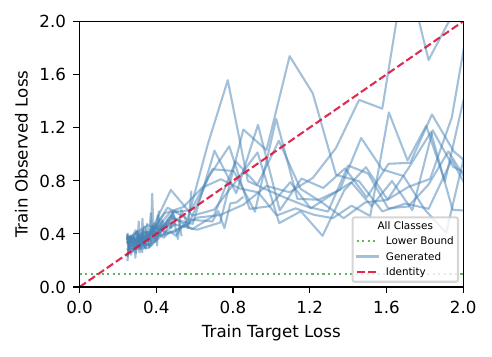}
        \caption{Losses comparison for class 2 - train}
        \label{fig:dihyfo2_mnist_obs_train2}
    \end{subfigure}
    \hfill
    % Second subfigure
    \begin{subfigure}[b]{0.48\textwidth}
        \centering
        \includegraphics[width=\textwidth]{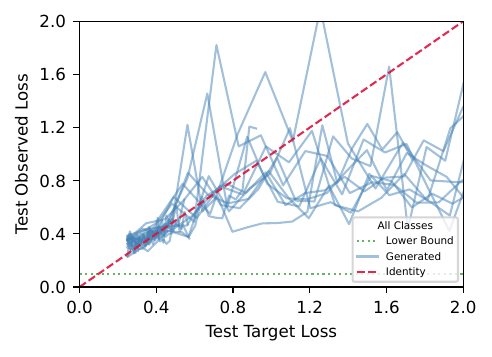}
        \caption{Losses comparison for class 3 - test}
        \label{fig:dihyfo2_mnist_obs_test2}
    \end{subfigure}
    \hfill
    
    \centering
    % First subfigure
    \begin{subfigure}[b]{0.48\textwidth}
        \centering
        \includegraphics[width=\textwidth]{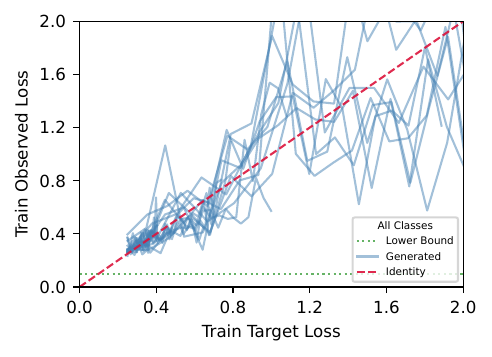}
        \caption{Losses comparison for class 3 - train}
        \label{fig:dihyfo2_mnist_obs_train3}
    \end{subfigure}
    \hfill
    % Second subfigure
    \begin{subfigure}[b]{0.48\textwidth}
        \centering
        \includegraphics[width=\textwidth]{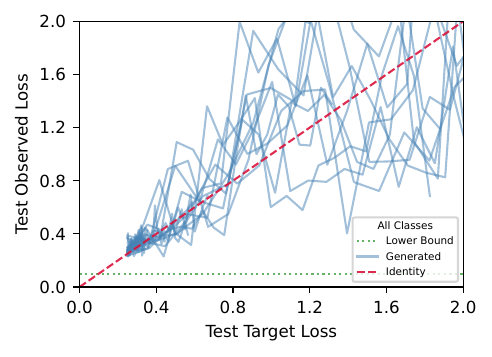}
        \caption{Losses comparison for class 3 - test}
        \label{fig:dihyfo2_mnist_obs_test3}
    \end{subfigure}
    \hfill
    
    \centering
    % First subfigure
    \begin{subfigure}[b]{0.48\textwidth}
        \centering
        \includegraphics[width=\textwidth]{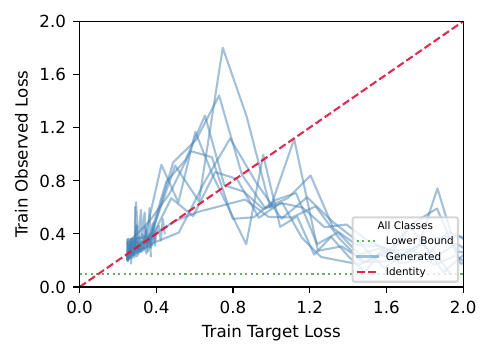}
        \caption{Losses comparison for class 9 (pivot) - train}
        \label{fig:dihyfo2_mnist_obs_train9}
    \end{subfigure}
    \hfill
    % Second subfigure
    \begin{subfigure}[b]{0.48\textwidth}
        \centering
        \includegraphics[width=\textwidth]{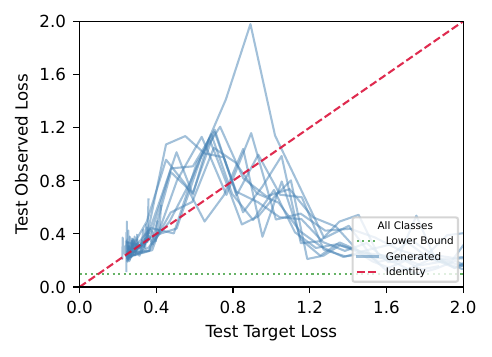}
        \caption{Losses comparison for class 9 (pivot) - test}
        \label{fig:dihyfo2_mnist_obs_test9}
    \end{subfigure}
    \hfill

    \caption{Comparison of losses obtained by the parameters generated with DiHyFo-2 for MNIST.}
    \label{fig:obs_des_dihyfo2_mnist}
\end{figure}

\begin{figure}[H]
\vspace{1in}
    \centering
    % First subfigure
    \begin{subfigure}[b]{0.472\textwidth}
        \centering
        \includegraphics[width=\textwidth]{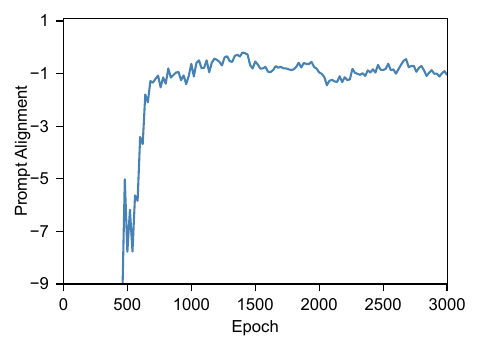}
        \caption{Prompt alignment for class 2}
        \label{fig:dihyfo2_mnist_c2r2}
    \end{subfigure}
    \hfill
    % Second subfigure
    \begin{subfigure}[b]{0.48\textwidth}
        \centering
        \includegraphics[width=\textwidth]{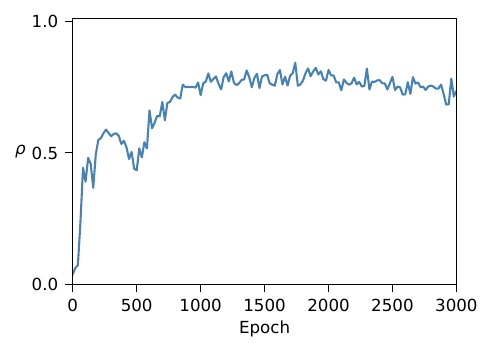}
        \caption{Correlation for class 2}
        \label{fig:dihyfo2_mnist_c2pearson_corr}
    \end{subfigure}
    \hfill

    \centering
    % First subfigure
    \begin{subfigure}[b]{0.472\textwidth}
        \centering
        \includegraphics[width=\textwidth]{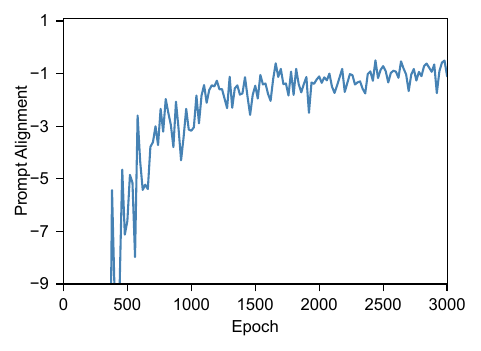}
        \caption{Prompt alignment for class 3}
        \label{fig:dihyfo2_mnist_c3r2}
    \end{subfigure}
    \hfill
    % Second subfigure
    \begin{subfigure}[b]{0.48\textwidth}
        \centering
        \includegraphics[width=\textwidth]{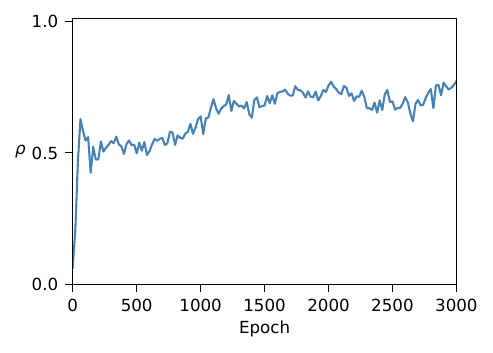}
        \caption{Correlation for class 3}
        \label{fig:dihyfo2_mnist_c3pearson_corr}
    \end{subfigure}
    \hfill
    
    \centering
    % First subfigure
    \begin{subfigure}[b]{0.472\textwidth}
        \centering
        \includegraphics[width=\textwidth]{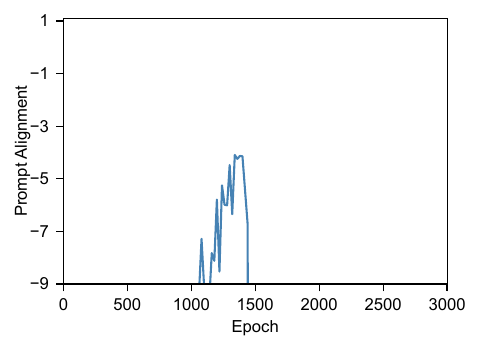}
        \caption{Prompt Alignment for class 9 (pivot)}
        \label{fig:dihyfo2_mnist_c9r2}
    \end{subfigure}
    \hfill
    % Second subfigure
    \begin{subfigure}[b]{0.48\textwidth}
        \centering
        \includegraphics[width=\textwidth]{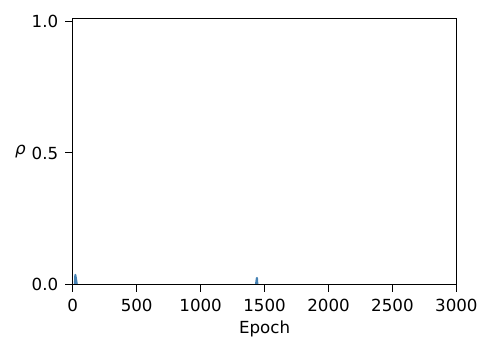}
        \caption{Correlation for class 9 (pivot)}
        \label{fig:dihyfo2_mnist_c9pearson_corr}
    \end{subfigure}
    \hfill

    \caption{Prompt alignment and correlation obtained by the parameters generated with DiHyFo-2 for MNIST.}
    \label{fig:pa_corr_dihyfo2_mnist}
\end{figure}

\newpage
\subsection{Additional hypernetworks, HyperForget, and DiHyFo details}
\label{subsec:add_hyperforget}

Consider a dataset $D=\{X,Y\}$ associated with a task $T$. In the classical deep learning framework the learnable parameters $\theta_F$ of a neural network $F(X; \theta_F)$ are obtained by solving the optimization problem in Equation \ref{eq:dl_optimize} \citep{chauhan2023brief}. 

The searching for the optimal configuration is performed within large search spaces formed by millions of potential parameters. The input samples $x \in X$ are passed through the layers of $F$ to obtain predictions $y^* \in Y^*$, later compared with the true labels $y \in Y$ using a loss function $L(Y, Y^*)$, which is optimized by updating parameters until convergence.

\begin{equation}
\label{eq:dl_optimize}
   \min_{\theta_F} F(X; \theta_F)
\end{equation}

\begin{figure}[H] 
\centering    
\includegraphics[width=0.6\textwidth]{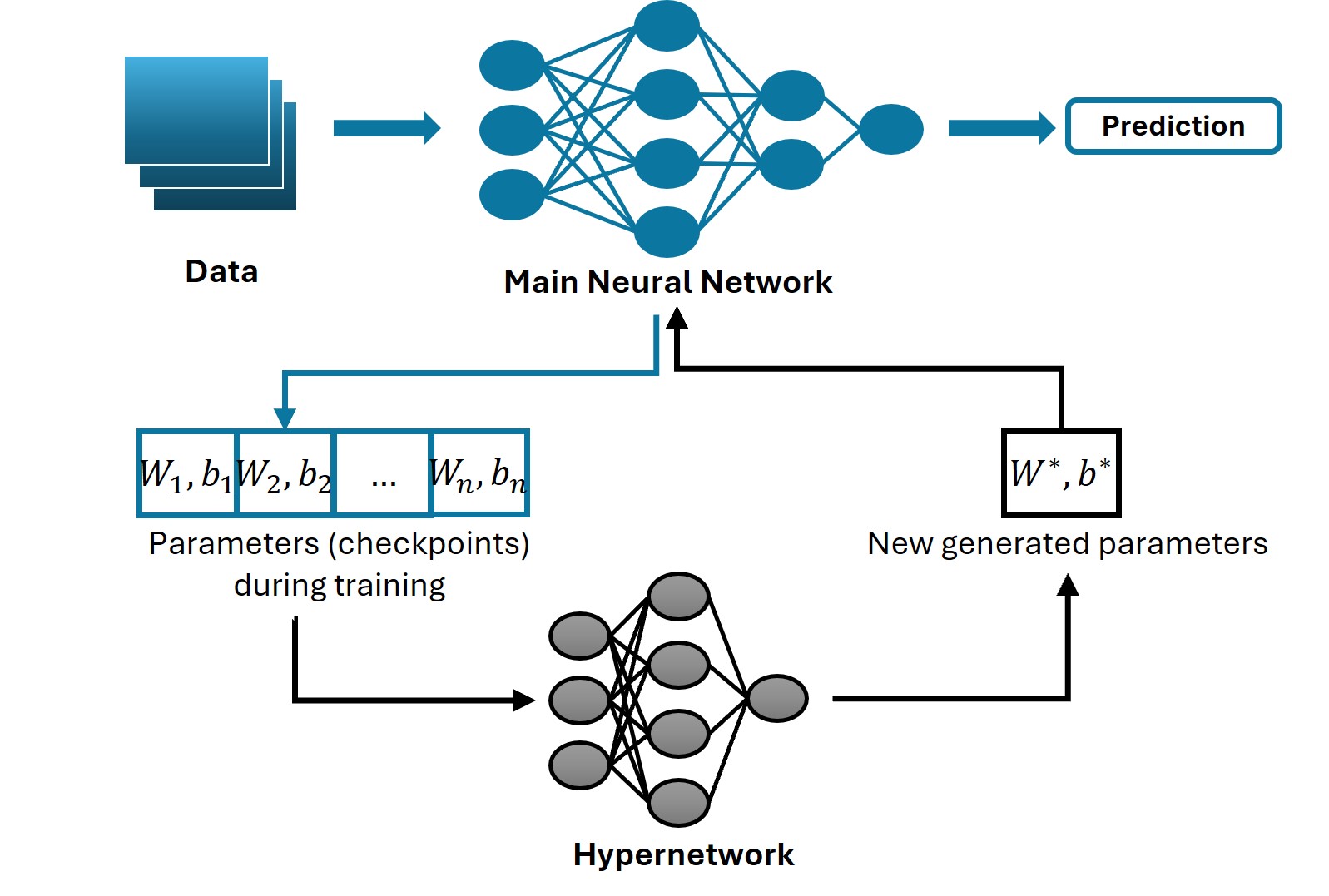}
\caption{Hypernetwork framework.}
\label{fig:HyperNetwork}
\end{figure}

Instead of directly optimizing the parameters of the main network, the hypernetwork framework, depicted in Figure \ref{fig:HyperNetwork}, uses a separate network to learn to generate the parameters for the main network. Both networks are usually trained in an end-to-end differentiable manner \citep{chauhan2023brief, ha2016hypernetworks}.

\begin{definition}[Hypernetwork]
\label{def:hypernetworks}
A neural network $G(C; \theta_G)$ is called a hypernetwork with learnable parameters $\theta_G$ and context input $C$ if its output are parameters for a second neural network $F(X;\theta_F)$ that solves a task $T$ with associated data $D=\{X,Y\}$, i.e. $\theta_F=G(C; \theta_G)$.
\end{definition}

The context input $C$ for the hypernetwork contains information about the structure of the parameters of the main network that enables learning to generate its parameters.

During the forward pass at training time, the parameters $\theta$ are generated by passing $C$ through $G$, and serve as input to $F$, which then process $x \in X$ and obtains predictions $y^* \in Y^*$. The loss $L(Y, Y^*)$ is then computed and during the backward pass, the error is back-propagated through $G$ with the gradients of $L$ computed with respect to $\theta_G$. Consequently, $\theta_G$ are optimized to generate the $\theta_F$ that best solves task $T$. This introduces the optimization problem in Equation \ref{eq:hyper_dl_optimize} \citep{chauhan2023brief}. 

At test time, new parameters $\theta_F^*$ can be sampled from the optimized hypernetwork and be used to make predictions with $F(X; \theta_F^*)$ on the test data. 

\begin{equation}
\label{eq:hyper_dl_optimize}
   \min_{\theta_G} F(X; G(C; \theta_G))
\end{equation}

For a classification task $T$ with training data $D=\{X,Y\}$, solved with a learning algorithm $F(X; \theta_F)$, the associated unlearning task with forget set $D_f \subseteq D$ and retain set $D_r=D \backslash D_f$ is solved by constructing an unlearned model $U(D,D_f,F)$ that is expected to perform equivalently or similarly to a model that has been trained without the forget set, $F(X \subseteq D\backslash D_f; \theta_F)$. To forget specific subsets of classes, the associated machine unlearning objective can be described by Equation \ref{eq:min_max_dihyfo}, where the first term increases the loss on the subset of classes to forget, $\mathcal{C}_\text{increase}$, reducing its influence, while the second minimizes the loss on the subset of classes to retain, $\mathcal{C}_\text{minimize}$, preserving model performance on it.

\begin{equation}
\label{eq:min_max_dihyfo}
     \max_{\theta} \bigg[ \sum_{c \in \mathcal{C}_\text{increase}} \mathbb{E}_{(x, y) \sim \mathcal{D}} \left[ \mathcal{L}_c(y, \hat{y}) \right] - \sum_{c \in \mathcal{C}_\text{minimize}} \mathbb{E}_{(x, y) \sim \mathcal{D}} \left[ \mathcal{L}_c(y, \hat{y}) \right] \bigg] \\
    \text{,} \quad \theta \in \Theta 
\end{equation}

\begin{figure}[H] 
\centering    
\includegraphics[width=0.3\textwidth]{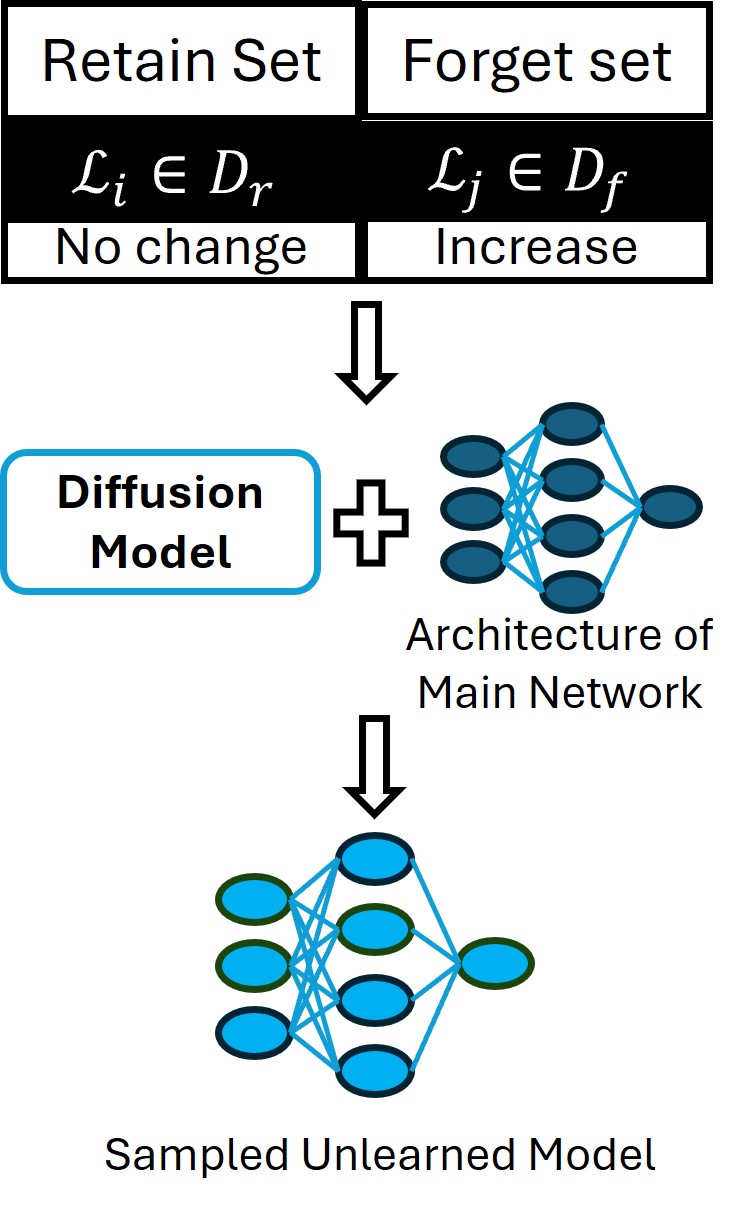}
\caption{DiHyFo process.}
\label{fig:unlearning_process_dihyfo}
\end{figure}

Consequently, the class unlearning task is closely related to control the influence of model parameters in the input-output interactions within the model. Instead of directly solving it, a hypernetwork can be conditioned on the specific class performance metrics, such as class losses, to construct a model capable of generating parameters that yield high-performance on $D_r$ while obtaining low-performance on $D_f$, effectively forgetting the specified classes, namely, a HyperForget model. Particularly, when a diffusion model is used as hypernetwork for the unlearning model, it receives the name of Diffusion HyperForget Network (DiHyFo). Figure \ref{fig:unlearning_process_dihyfo} illustrates this procedure.

Notably, unlearned models obtained using HyperForget can be interpreted and evaluated using the two interpretations of the definition of unlearning, either as an approximation to exact unlearning on the parameter space or in the output space \citep{nguyen2022survey, Bourtoule2021MachineUnlearning, ginart2019making, goel2022evaluating}.

\begin{figure}[H]
    \centering
    % First subfigure
    \begin{subfigure}[b]{0.6\textwidth}
        \centering
        \includegraphics[width=\textwidth]{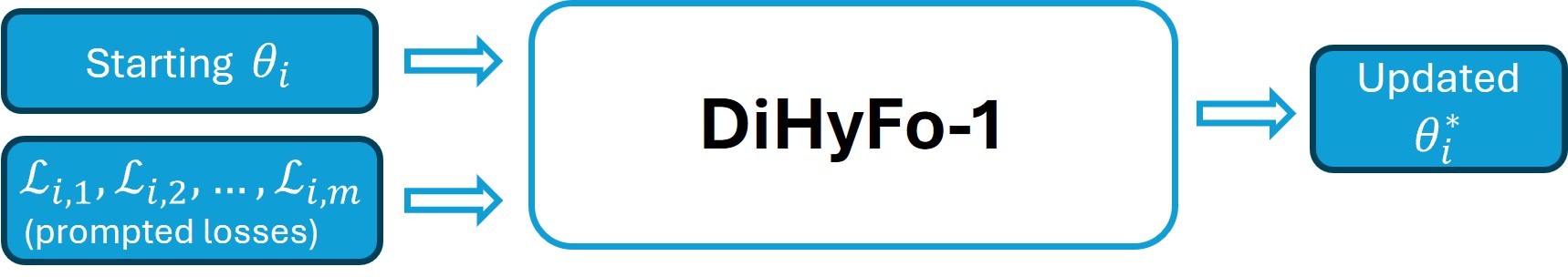}
        \caption{DiHyFo-1}
        \label{fig:dyhifo1_simple}
    \end{subfigure}
    
    \vspace{0.6cm} % Add vertical space between the subfigures
    
    % Second subfigure
    \begin{subfigure}[b]{0.6\textwidth}
        \centering
        \includegraphics[width=\textwidth]{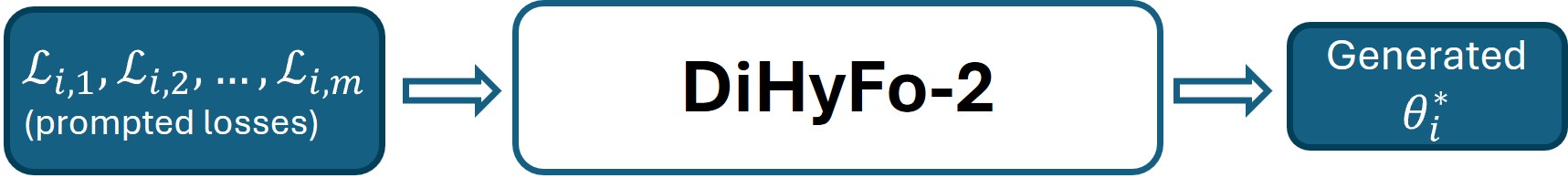}
        \caption{DiHyFo-2}
        \label{fig:dyhifo2_simple}
    \end{subfigure}
    
    \caption{Two implementations of DiHyFo.}
    \label{fig:dihyfos_approaches}
\end{figure}

We have presented two mechanisms for constructing a DiHyFo model, depicted in Figure \ref{fig:dihyfos_approaches}. DiHyFo-1 is built on the learning-to-learn framework of \cite{Peebles2022LearningTL}, extending it to generate parameters conditioned on class losses and to be able to learn to deoptimize, i.e., learning-to-forget. Conversely, DiHyFo-2 uses a diffusion model directly conditioned on the desired class losses.

DiHyFo-1's goal is to predict the distribution of updated parameters that achieve the target loss for each class, as detailed in Equation \ref{eq:gpt_gen_task}. The model's architecture is illustrated in Figure \ref{fig:dyhifo1_full}.

\begin{equation}
\label{eq:gpt_gen_task}
p_G(\theta^*\mid \theta, \mathcal{L}_1^*, ..., \mathcal{L}_m^*, \mathcal{L}_1, ..., \mathcal{L}_m)    
\end{equation}

The training objective is to minimize the MSE between the original future parameters and the parameters generated by the DiT so that it learns to generate parameters that closely match the original updated parameters conditioned on the input losses, Equation \ref{eq:mse_dihyfo}.

\begin{equation}
\label{eq:mse_dihyfo}
 \mathfrak{L}(G)=\mathbb{E}(\|\theta^* - G(\theta_t^*,\theta, \mathcal{L}_1^*,...,\mathcal{L}_m^*, \mathcal{L}_1,...,\mathcal{L}_m, t)\|_2^2)
\end{equation}

The task of DiHyFo-2 is to predict the distribution of parameters that achieve the desired losses conditioned directly on them, reducing all previous expressions to this simpler conditioning.

\subsection{Additional details on evaluation metrics}
\label{subsec:evaluation_metrics}

Each model must be evaluated on two key aspects. First, their generative properties as DiHy need to be assessed. This involves verifying whether the models can generate appropriate parameters for the main network, enabling it to approximate targeted performance levels on the classification task for each class. 

Both DiHyFo models use MSE as training metric, and we analyze the corresponding learning curves to understand the models' learning behavior and convergence, and identify issues like overfitting or underfitting.

During evaluation, the model is prompted with a value close to the best loss found in the training dataset. It is noted in \cite{Peebles2022LearningTL} that using a value slightly above or below the best loss can sometimes yield better results for certain tasks. To evaluate whether the generated parameters effectively approximate prompted loss values, we use the prompt alignment metric defined by \citep{Peebles2022LearningTL} as the $R^2$ score between the obtained loss and the target loss in Equation \ref{eq:r2_score}, averaged over the batch size.

\begin{equation}
\label{eq:r2_score}
R^2 = 1 - \frac{\sum_{i=1}^n (\mathcal{L}^*_{i} - \hat{\mathcal{L}}^*_{i})^2}{\sum_{i=1}^n (\mathcal{L}^*_{i} - \eta)^2 + \epsilon}
\end{equation}

where \(\epsilon\) is a small constant to avoid division by zero, and \(\eta\) is the mean of the targets:

\begin{equation}
\eta\ = \frac{1}{n} \sum_{i=1}^n \mathcal{L}^*_{i}
\end{equation}

A score close to 1 indicates that the obtained loss closely aligns with the target loss. Negative values suggest worse alignment than the mean of the target values. We compute this score separately for each class to assess the model's ability to control losses across different classes simultaneously. And it is computed over 20 regularly-sampled prompts and averaged over multiple neural networks sampled with each DiHyFo, using randomly-initialized input parameters.

This metric is effective for evaluating conditional generative tasks, such as those in the learning-to-learn framework (\cite{Peebles2022LearningTL}), as it measures how well the model aligns the observed losses with the target losses in terms of both direction and magnitude. However, our training datasets exhibit bidirectional loss movements, which can affect how the model learns to align the losses. In our experiments, we found that observed losses generally aligned correctly with the direction of the prompted losses, but not always in magnitude, and sometimes some generated parameters have drastic undesired performances that significantly drop the prompt alignment in comparison with the common behavior of the generated parameters. Thus, we compute the Pearson correlation between observed and target losses to assess if the model is accurately tracking the target, providing a complementary metric to the prompt alignment score.

Additionally, we directly compare the observed versus target losses by plotting them 
along the identity line (indicating perfect alignment) and include a lower performance bound—often set to the median or average of losses from checkpoint collection—as a reference for high performance.

The second aspect to evaluate is the DiHyFo models' ability to sample unlearned networks. We assess the unlearning properties of networks sampled with each DiHyFo by comparing them to a network retrained from scratch without the forget targets.

We compare the individually accuracy of each unlearned model and the retrained model on the retain, forget, and complete test sets. Additionally, we calculate the MIA score for each model to assess the likelihood that an adversary could infer information about the forget set from the unlearned model. Ideally, the unlearned model should have MIA scores similar to the retrained model. To compute the MIA score, we follow the implementation of \cite{chundawat2023can, foster2024fast}, as in Figure \ref{fig:MIA_algorithm}. The model being evaluated generates prediction probabilities for the retain and test sets, these probabilities are used to calculate entropy for each set, and the resulting entropies serve as features for a logistic regression with labels indicating the origin dataset of each instance. The logistic regression is trained and then used to predict the membership status of instances in the forget set. The average of these predictions provides the MIA score, reflecting how effectively it can infer whether the forget data was part of the training set of the model being evaluated.

\begin{figure}[h!]
    \centering
    \begin{minipage}{0.9\textwidth}
        \begin{algorithmic}[1]
            \State \textbf{Input:} Model $F$, Datasets $D_r$ and $D$ (retain and test sets)
            \State $P_r \gets F(D_r)$
            \State $P_t \gets F(D)$
            \State $X_r \gets [\mathcal{H}(P_r), \mathcal{H}(P_t)]$ \Comment{$\mathcal{H}$ is the entropy measure}
            \State $Y_r \gets [1 \text{ (retain)}, 0 \text{ (test)}]$
            \State LR $\gets$ Train Logistic Regression with $(X_r, Y_r)$
            \State $\hat{Y}_f \gets \text{LR}(\mathcal{H}(P_r))$
            \State MIA $\gets \frac{1}{n_f} \sum_{i=1}^{n_f} \hat{Y}_{fi}$
            \State \textbf{Output:} MIA
        \end{algorithmic}
    \end{minipage}
    \caption{Pseudocode for computing MIA score.}
    \label{fig:MIA_algorithm}
\end{figure}

As we employ a retrained model from scratch without the forget set as baseline, we would like the unlearned models sampled with each DiHyFo to behave as close as possible to the retrained model. Thus, to compare each unlearned model's output space against the retrained model, we measure the overlap in their predictions on each set. For parameter space comparison we compute an \textit{unlearning score} $\varphi$ using the Jensen-Shannon Divergence (JSD) \citep{chundawat2023can, nguyen2022survey, foster2024informationtheoreticapproachmachine}, which presents a symmetric measure constructed using the Kullback–Leibler divergence ($KL$) between two probability distributions, $P$ and $Q$. For the unlearning score, the JSD is computed between the softmax of each pair of unlearned models being compared across the entire dataset, Equation \ref{eq:js_divergence}, in our case the sampled unlearned model using a DiHyFo and the retrained model. The closer the score value to 1, the greater the similarity in behavior between the models for the same inputs.

\begin{equation}
S(\mathbf{z}_i) = \frac{e^{z_i}}{\sum_{j=1}^K e^{z_j}} \nonumber
\end{equation}

 \begin{align}
\label{eq:js_divergence}
\text{JSD}(P \parallel Q) &= \frac{1}{2} \text{KL}\left(P \parallel M\right) + \frac{1}{2} \text{KL}\left(Q \parallel M\right) \\
M &= \frac{P + Q}{2} \nonumber
\end{align}

% UnLearning Score
\begin{equation}
\varphi = 1 - \text{JSD}(S(\mathbf{z}_{\text{model}_1}) \parallel S(\mathbf{z}_{\text{model}_2}))
\end{equation}

All these metrics provide a thorough assessment of each DiHyFo model's ability to generate parameters conditioned on class losses and their suitability for unlearning tasks, which is our primary goal.

\subsection{Additional details on datasets generation}
\label{subsec:datasets_generation}

\begin{figure}[H]
    \centering
    \includegraphics[width=0.7\textwidth]{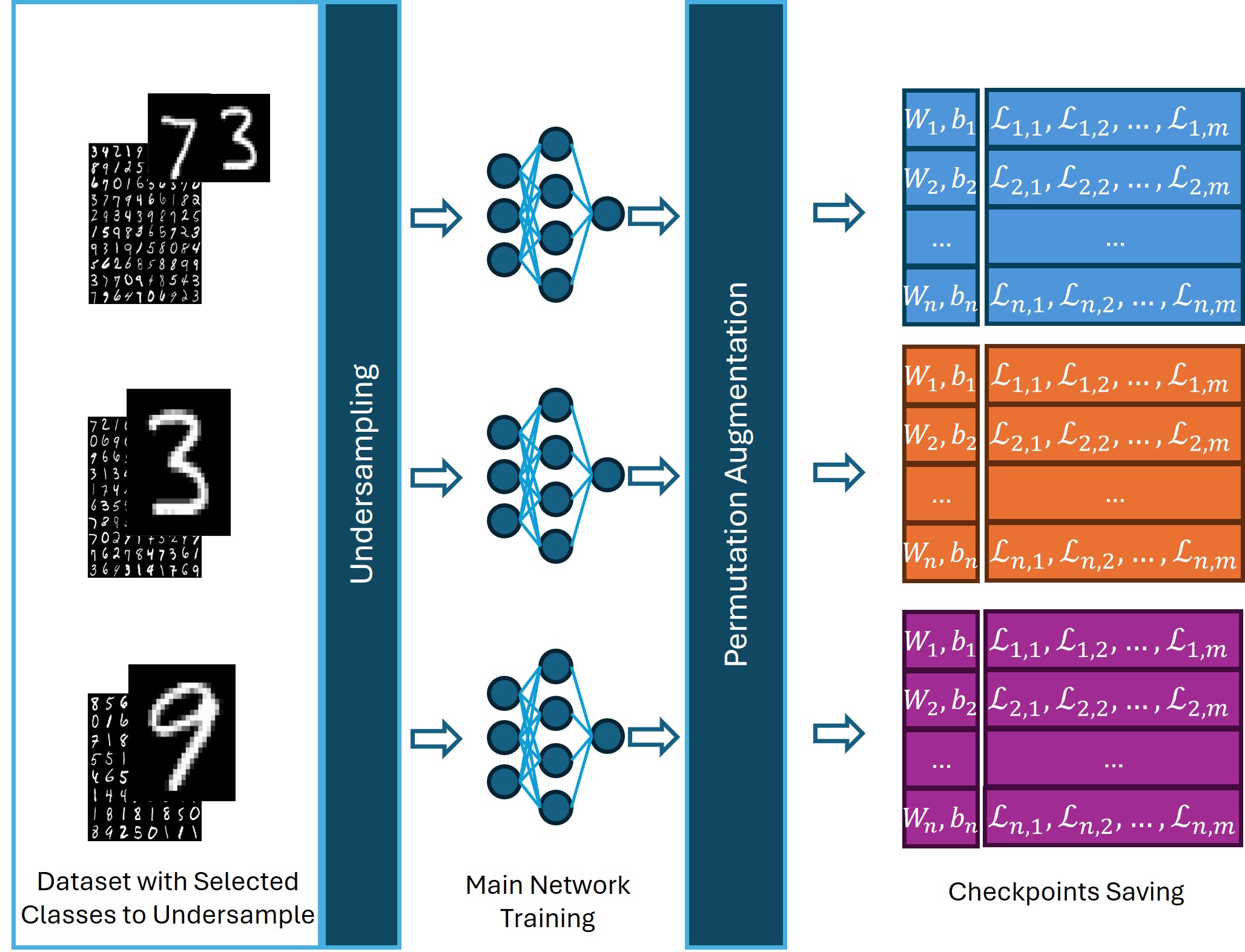}
    %\caption[Checkpoints Collection for the Optimization Process]{Checkpoints collection for the Optimization Process}
    \caption{Checkpoints collection for the optimization process.}
    \label{fig:data_gen_1}
\end{figure}

Checkpoints of parameters and associated class losses are collected during multiple training runs of an MLP on MNIST. Since a simple MLP achieves good results on MNIST early in training, we randomly select a subset of classes to undersample in each run to capture a broader range of loss values. During the MLPs training, a checkpoint is randomly selected and evaluated for potential saving. Permutation augmentation is applied as in \citep{Peebles2022LearningTL}. The process is illustrated in Figure \ref{fig:data_gen_1}.

As SGD is an unconstrained optimizer, we make use of heuristics to track the evolution of class losses in a constrained manner, saving checkpoints that include parameters that perform well for some classes and bad for others simultaneously.

Initially, we create bins corresponding to different loss levels and establish a maximum of examples per bin to prevent over-collecting certain loss levels while allowing for more collection on non-frequent loss levels.  As the wide range of possible loss values across multiple classes leads to a combinatorial explosion of potential loss-level combinations, we consider a simplified scenario where for a classification task with $m$ classes we use $r$ classes as pivots ($r<m$). We only consider checkpoints when the model achieves high performance on the pivots, using an accuracy threshold $\gamma=80$. Lowering this threshold increases variability in pivots and expands the combinatorial possibilities. The remaining $m-r$ classes are allowed to vary across the full range of possible loss values. During each training run, if a randomly selected iteration has pivots' performance over  $\gamma$ and the corresponding bin for remaining classes is not full, the checkpoint is saved. In this setup, the forget set can be any subset of the  $m-r$ classes, while the retain set must always include the pivots.

For more complicated scenarios, we relax the constrains, increasing the number of classes and using fewer pivots or none at all. Instead of using bins to categorize class-level losses, we implement checkpoint-saving conditions to balance the need for diverse loss examples with computational efficiency. 

We prioritize capturing high and low performing parameter updates simultaneously, apply diverse undersampling rates and increase the random selection rate during early training epochs, while for later epochs the selection rate is reduced to focus on capturing significant shifts in loss. Also, in some runs we randomly select checkpoints across the training to ensure the models have access to examples of both the general loss evolution process and the particular moments we are interested in for the forgetting task. This checkpoint collection approach has proven effective for conditional parameter generation using DiHy \citep{li2024text, wang2024networkdiffusion}, and ensures our dataset includes key learning moments while avoiding redundancy. Figure \ref{fig:checkpoint_algorithm}  summarizes this checkpoint-saving strategy.

\begin{figure}[H]
    \centering
    \includegraphics[width=0.8\textwidth]{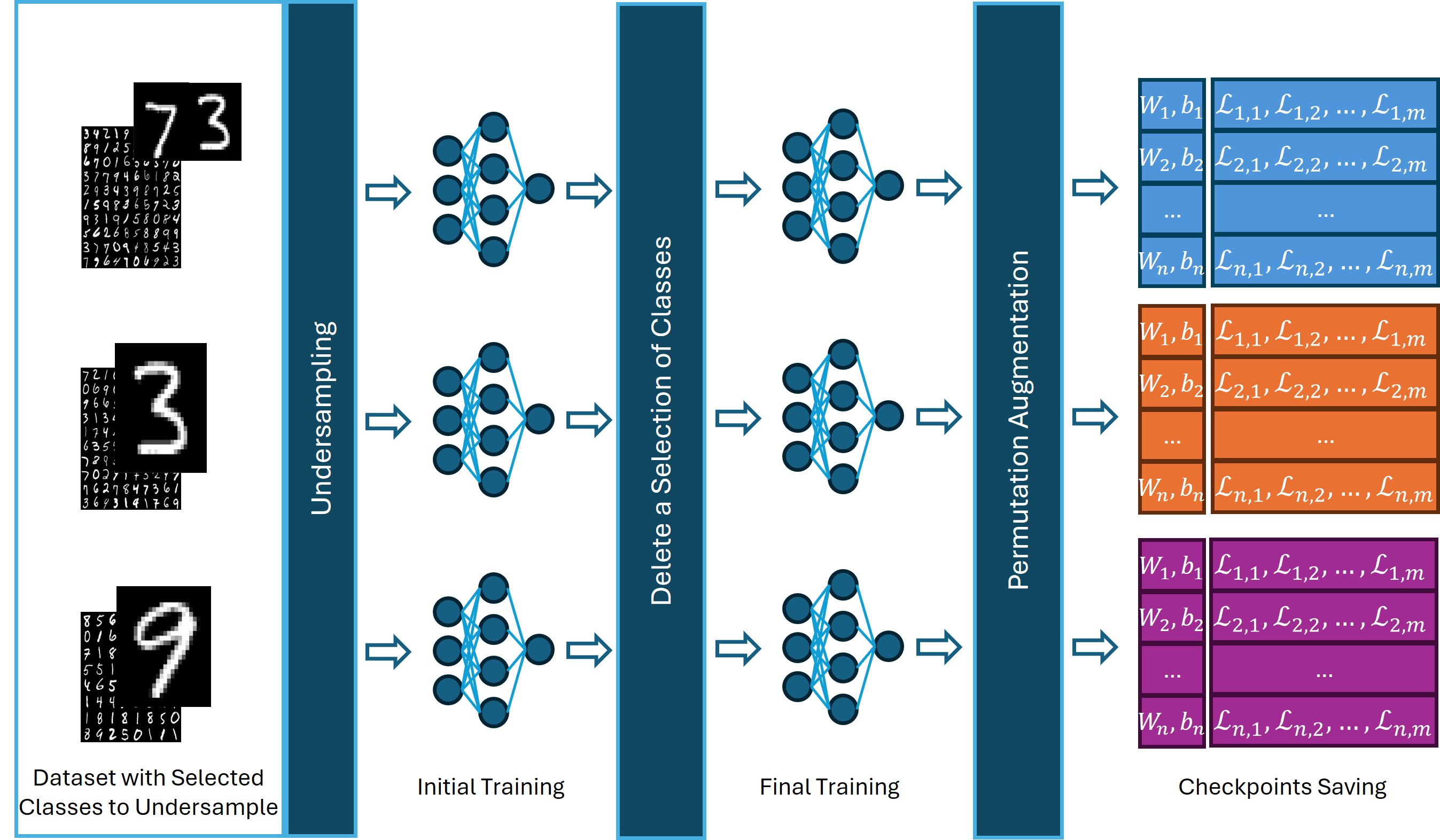}
    %\caption[Checkpoints Collection for the Optimization Process]{Checkpoints collection for the Optimization Process}
    \caption{Checkpoints collection for de-optimization process.}
    \label{fig:data_gen_2}
\end{figure}

In the case of DiHyFo-1, it needs samples that enable it to learn bidirectional movements in class losses, either by reversing the order of checkpoint loading during DiT training or by saving checkpoints from a process with incremental losses. We chose the latter approach so that the model can learn directly from a forgetting process. To collect examples from the forgetting process, we follow a similar training procedure as described before, but at a certain point during training, we randomly delete a selection of classes to capture the associated increase in losses. This is similar to forgetting by fine-tuning the main network without the classes to be forgotten, Figure \ref{fig:data_gen_2}.

DiHyFo-2 does not requires to see a process of increments in class losses, just a variety of combinations of losses across classes, which can be done by collecting a vast amount of checkpoints from runs with different configurations.

\begin{figure}[h!]
    \centering
    \begin{minipage}{0.9\textwidth}
        \begin{algorithmic}[1]
            
            \State \textbf{Input:} $\gamma$ (performance threshold), $F$ (main network), $D_T$ (task data), $\beta$ (checking threshold), $bin_1,...,bin_r$ (loss bins), $b_1,..., b_r$ (max bin sizes), $Y_p$ (pivots), max\_checkpoints

            \State $D_1 \gets [X_{\text{training}}, Y_{\text{training}}]$ from $D$
            \State $D_2 \gets [X_{\text{test}}, Y_{\text{test}}]$ from $D$

            \State $J \gets \text{Random sample from unique}(Y_{\text{training}})$
            \State $D_1 \gets \text{subsample}(D_1, J)$
            \State $n_{\text{checkpoints}} \gets 0$
            
            \For{epoch in $N_{\text{epochs}}$}
                \If{$n_{\text{checkpoints}} < \text{max\_checkpoints}$}
                    \State training Step $F(D_1)$
                    \State $\mathbf{L_1,...,L_m} \gets \text{Compute class losses of } F(D_1) \text{ on } D_2$
                    \If{$\beta < \text{sampled random value}$}
                        \If{every $L_j$ associated with $Y_p > \gamma$}
                            \If{$L_u$ in $bin_u$, not $L_u$ in $Y_p$}
                                \If{count($bin_u$) < $b_u$}
                                    \State save checkpoint
                                    \State $n_{\text{checkpoints}} \gets n_{\text{checkpoints}}+1$
                                \EndIf
                            \EndIf
                        \EndIf
                    \EndIf
                \EndIf
            \EndFor
        \end{algorithmic}
    \end{minipage}
    \caption{Pseudocode for collecting checkpoints with bins.}
    \label{fig:checkpoint_algorithm}
\end{figure}

\subsection{Some experimental observations on G.pt}
\label{subsec:gpt_observations}

Our experiments suggest that, while G.pt can learn to generate parameters using loss, prediction error, or accuracy, it is generally easier to learn with loss. Losses values typically span a narrower range than prediction error or accuracy, making them more sensitive to parameter changes. Therefore, we decided to use loss as our conditional performance metric.

\begin{figure}[H]
    \centering
    % First subfigure: Train
    \begin{subfigure}[b]{0.48\textwidth}
        \centering
        \includegraphics[width=\textwidth]{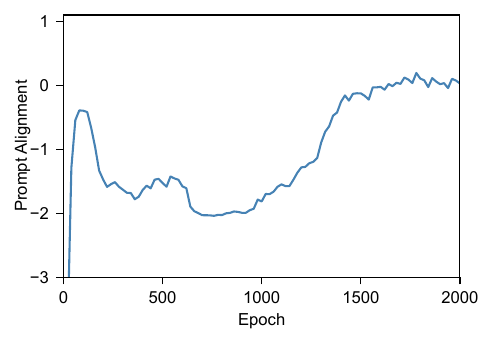}
        \caption{Train}
        \label{fig:gpt_error_trainr2}
    \end{subfigure}
    \hfill
    % Second subfigure: Test
    \begin{subfigure}[b]{0.48\textwidth}
        \centering
        \includegraphics[width=\textwidth]{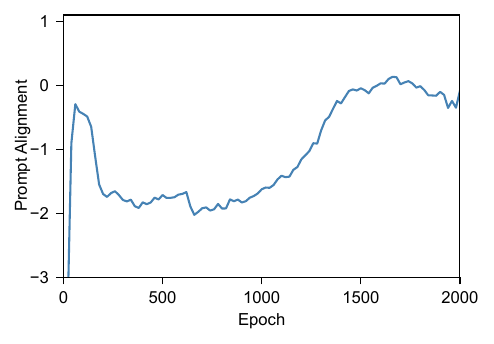}
        \caption{Test}
        \label{fig:gpt_error_testr2}
    \end{subfigure}
    
    \caption{Evolution of G.pt prompt alignment with prediction error as metric.}
    \label{fig:gpt_results_error}
\end{figure}

\begin{figure}[H]
    \centering

    % First row: First and Second subfigure
    \begin{subfigure}[b]{0.48\textwidth}
        \centering
        \includegraphics[width=\textwidth]{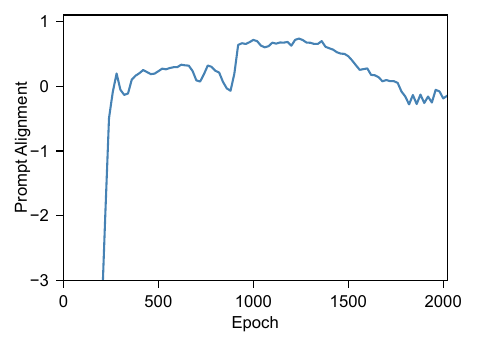}
        \caption{Train prompt alignment}
        \label{fig:gpt_trainr2}
    \end{subfigure}
    \hfill
    \begin{subfigure}[b]{0.48\textwidth}
        \centering
        \includegraphics[width=\textwidth]{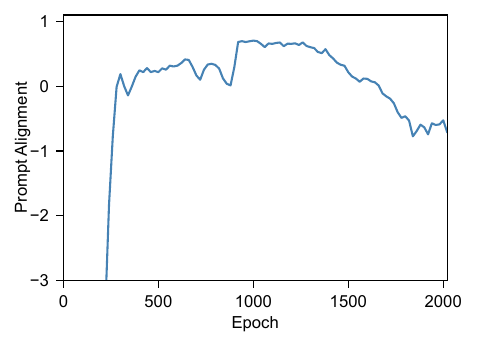}
        \caption{Test prompt alignment}
        \label{fig:gpt_testr2}
    \end{subfigure}

    \vspace{0.5cm} % Add some vertical space between rows

    % Second row: Third and Fourth subfigure
    \begin{subfigure}[b]{0.48\textwidth}
        \centering
        \includegraphics[width=\textwidth]{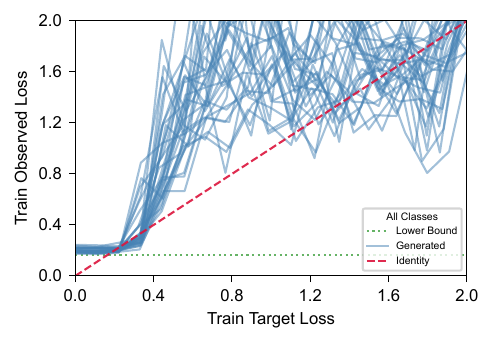}
        \caption{Train observed vs target loss}
        \label{fig:gpt_loss_orig_obs_train}
    \end{subfigure}
    \hfill
    \begin{subfigure}[b]{0.48\textwidth}
        \centering
        \includegraphics[width=\textwidth]{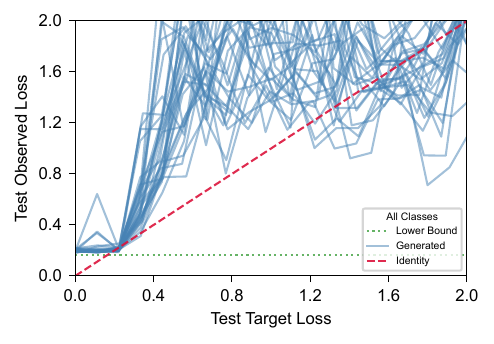}
        \caption{Test observed vs target loss}
        \label{fig:gpt_loss_orig_obs_test}
    \end{subfigure}

    \caption{G.pt training results.}
    \label{fig:gpt_results}
\end{figure}

CIFAR-10 experimental results showed in \cite{Peebles2022LearningTL} used prediction error as a conditional metric with positive results. However, for this particular experiment it was observed that the range of prediction error values obtained by the target model is contained in a relatively narrower range than usual. When comparing performance across different metrics on the same set, as shown in Figure \ref{fig:gpt_results_error}, the instances with more samples and a narrower value range, typically loss, performed better.

\begin{figure}[H]
\vspace{-0.05in}
    \centering
    % First subfigure: Train
    \begin{subfigure}[b]{0.48\textwidth}
        \centering
        \includegraphics[width=\textwidth]{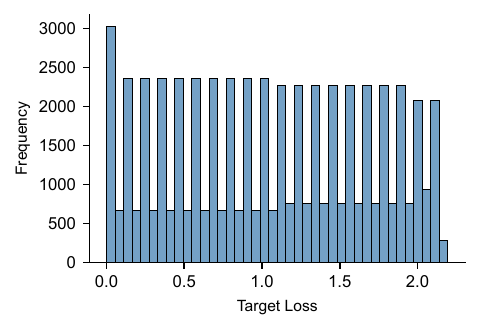}
        \caption[Distribution of Target Losses]{Target losses}
        \label{fig:gpt_loss_frec_bins40}
    \end{subfigure}
    \hfill
    % Second subfigure: Test
    \begin{subfigure}[b]{0.48\textwidth}
        \centering
        \includegraphics[width=\textwidth]{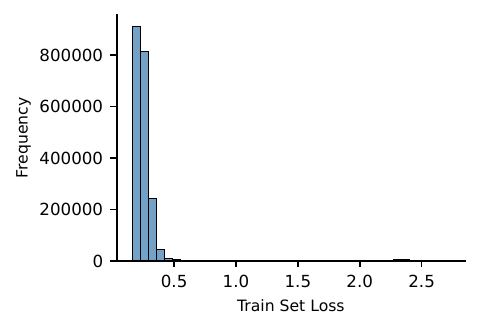}
        \caption[Distribution of Losses in Train Set]{Losses in train set}
        \label{fig:train_set_40}
    \end{subfigure}
    
    \caption{Comparison of distribution of test losses and losses in train set for G.pt.}
    \label{fig:gpt_loss_frec_bins}
\end{figure}

\begin{figure}[H]
\vspace{-0.2in}
    \centering
    % First subfigure: Train
    \begin{subfigure}[b]{0.48\textwidth}
        \centering
        \includegraphics[width=\textwidth]{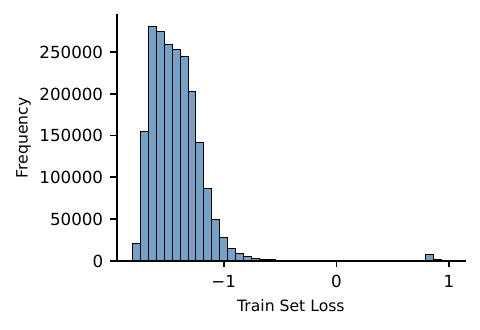}
        \caption{Losses in train set}
        \label{fig:train_set_40_log}
    \end{subfigure}
    \hfill
    % Second subfigure: Test
    \begin{subfigure}[b]{0.48\textwidth}
        \centering
        \includegraphics[width=\textwidth]{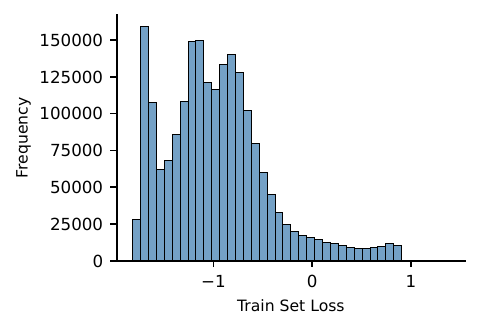}
        \caption{Losses in resampled train set}
        \label{fig:balanced_train_set_40_log}
    \end{subfigure}
    
    \caption{Comparison of distribution of losses in train set and resampled train set for G.pt in log-scale.}
    \label{fig:gpt_loss_frec_bins2}
\end{figure}

\begin{figure}[H]
\vspace{-0.2in}
    \centering
    % First subfigure: Train
    \begin{subfigure}[b]{0.48\textwidth}
        \centering
        \includegraphics[width=\textwidth]{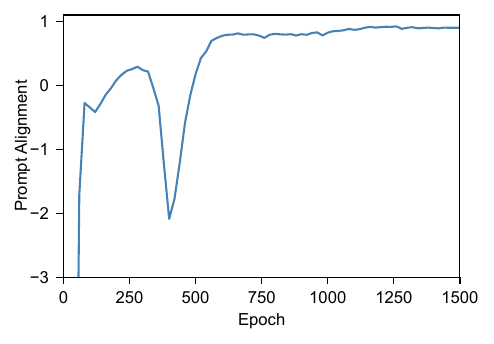}
        \caption{Prompt alignment - train}
        \label{fig:gpt_loss_bal_trainr2}
    \end{subfigure}
    \hfill
    % Second subfigure: Test
    \begin{subfigure}[b]{0.48\textwidth}
        \centering
        \includegraphics[width=\textwidth]{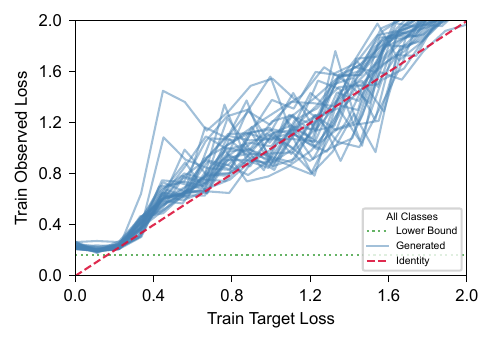}
        \caption{Observed vs target loss - train}
        \label{fig:gpt_error_testr2_balanced}
    \end{subfigure}
    
    \caption{Behavior of G.pt with re-balanced loss data.}
    \label{fig:gpt_results_bal}
\end{figure}

Additionally, Figure \ref{fig:gpt_loss_frec_bins} highlights a disparity between the distribution of losses found during testing and those in the training datasets used to produce the results in Figure \ref{fig:gpt_results}. During testing, metric values tend to be more uniformly distributed, thus, effective learning requires varied examples across the entire testing range. By resampling the training set to cover a wider range of losses, Figure \ref{fig:gpt_loss_frec_bins2}, G.pt's performance improves significantly, as shown in Figure \ref{fig:gpt_results_bal}. Indeed, experimental observations indicate that G.pt, while capable of generating neural network parameters for diverse losses, relies heavily on well-composed training data to perform effectively within the target metric space.

\begin{figure}[H]
%\vspace{-80pt}
    \centering
    % First subfigure
    \begin{subfigure}[b]{0.48\textwidth}
        \centering
        \includegraphics[width=\textwidth]{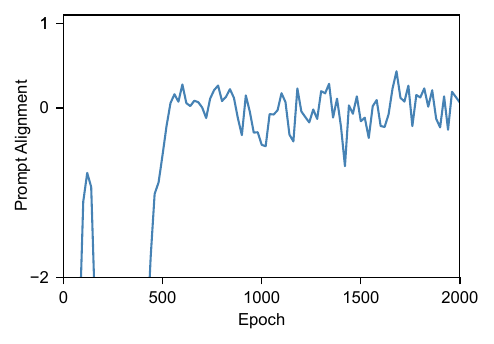}
        \caption{Prompt alignment - train.}
        \label{fig:gpt_1class_trainr2}
    \end{subfigure}
    \hfill
    % Second subfigure
    \begin{subfigure}[b]{0.48\textwidth}
        \centering
        \includegraphics[width=\textwidth]{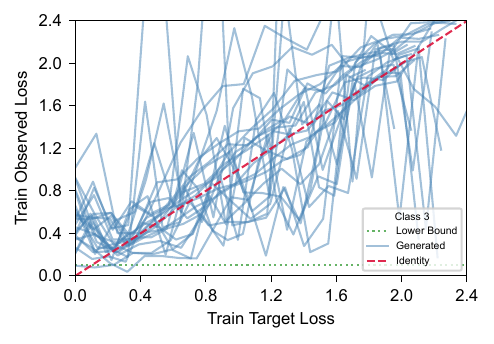}
        \caption{Loss comparison - train.}
        \label{fig:gpt_1class_obs_train}
    \end{subfigure}
    \hfill
    % Second subfigure
    \begin{subfigure}[b]{0.48\textwidth}
        \centering
        \includegraphics[width=\textwidth]{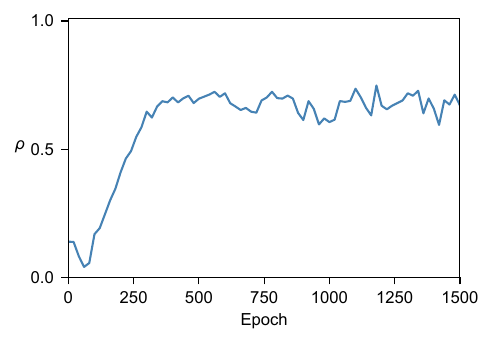}
        \caption{Correlation - train.}
        \label{fig:gpt_1class_trainpearson_corr}
    \end{subfigure}
    \hfill
    \centering
    % First subfigure
    \begin{subfigure}[b]{0.48\textwidth}
        \centering
        \includegraphics[width=\textwidth]{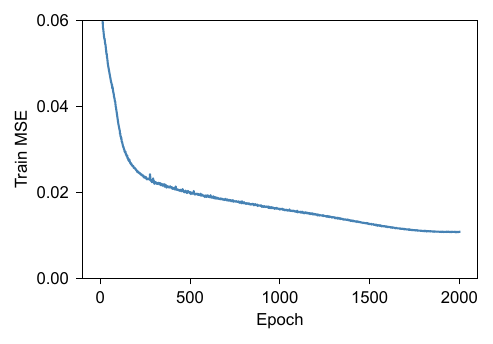}
        \caption{Training learning curve.}
        \label{fig:gpt_1class_train_loss}
    \end{subfigure}
    \hfill

    \caption{Behavior of G.pt when trained conditioned on one class loss.}
    \label{fig:gpt_1class}
\end{figure}

On the other hand, we evaluated G.pt capabilities when conditioned on individual classes. Figure \ref{fig:gpt_1class} illustrates that while G.pt had potential to learn from conditioning on one class loss, it still needs additional considerations to be able to learn from multiple classes simultaneously and more to learn to forget.

\end{document}